\documentclass{article}
\PassOptionsToPackage{numbers,compress}{natbib}
\usepackage[preprint]{neurips_2026}

\usepackage[utf8]{inputenc}
\usepackage[T1]{fontenc}
\usepackage{amsmath}
\usepackage{amssymb}
\usepackage{amsfonts}
\usepackage{nicefrac}
\usepackage{graphicx}
\usepackage{booktabs}
\usepackage{microtype}
\usepackage{xcolor}
\usepackage[normalem]{ulem}
\usepackage{listings}
\usepackage{float}      
\usepackage{placeins}   
\definecolor{linknavy}{rgb}{0.05,0.25,0.55}
\definecolor{citegreen}{rgb}{0.05,0.40,0.25}
\usepackage[colorlinks=true,linkcolor=linknavy,citecolor=citegreen,
            urlcolor=linknavy,breaklinks=true]{hyperref}

\definecolor{jsonkey}{HTML}{0B5394}
\definecolor{jsontype}{HTML}{707880}
\definecolor{jsonrule}{HTML}{D8DEE4}
\definecolor{jsonback}{HTML}{F7F8FA}

\lstdefinelanguage{jsonschema}{
  morestring=[b]",
  moredelim=[s][\color{jsontype}\itshape]{<}{>},
  showstringspaces=false,
  sensitive=false,
}

\lstdefinestyle{jsonstyle}{
  language=jsonschema,
  basicstyle=\ttfamily\footnotesize,
  stringstyle=\color{jsonkey},
  frame=single,
  rulecolor=\color{jsonrule},
  backgroundcolor=\color{jsonback},
  framesep=4pt,
  framexleftmargin=3pt,
  xleftmargin=3pt,
  xrightmargin=3pt,
  aboveskip=\medskipamount,
  belowskip=\medskipamount,
  columns=fullflexible,
  keepspaces=true,
  breaklines=false,
}
\lstdefinelanguage{prompttext}{
  morestring=[b]",
  showstringspaces=false,
  sensitive=false,
}

\lstdefinestyle{promptstyle}{
  language=prompttext,
  basicstyle=\ttfamily\scriptsize,
  stringstyle=\color{jsonkey},
  frame=single,
  rulecolor=\color{jsonrule},
  backgroundcolor=\color{jsonback},
  framesep=4pt,
  framexleftmargin=3pt,
  xleftmargin=3pt,
  xrightmargin=3pt,
  aboveskip=\medskipamount,
  belowskip=\medskipamount,
  columns=fullflexible,
  keepspaces=true,
  breaklines=false,
}

\newcommand{\beginsupplement}{%
  \setcounter{figure}{0}\renewcommand{\thefigure}{S\arabic{figure}}%
  \renewcommand{\theHfigure}{S\arabic{figure}}%
  \setcounter{table}{0}\renewcommand{\thetable}{S\arabic{table}}%
  \renewcommand{\theHtable}{S\arabic{table}}}

\begin{document}
\raggedbottom
\title{Confidence scores for gated automation of large language models}

\author{%
  Luiz G. A. Alves \\
  Egen AI \\
  Chicago, IL, USA \\
  \texttt{luizgustavo.deandradealves@egen.ai} \\
  \And
  Aaron Vermeersch \\
  Egen AI \\
  Chicago, IL, USA \\
  \texttt{aaron.vermeersch@egen.ai} \\
}

\maketitle

\begin{abstract}
AI agents built on large language models (LLMs) demonstrate strong capabilities in automated decision-making. However, LLMs may produce incorrect answers with high confidence, limiting their applicability in decisions that affect individuals and organizations. Reliable confidence scores are necessary to gate such decisions, and existing estimates are frequently unavailable or too coarse to support effective gating. This work uses supervised learning to train models to predict the correctness of LLM-generated answers, enabling automated confident responses and routing uncertain ones for human review. A linear confidence score head is fitted to the final-layer hidden state of the frozen LLM backbone, and text-only metamodels are trained as an alternative for closed models. All results come from a single 2B-parameter backbone, on graded answers that are about 98\% SQuAD and the remainder TruthfulQA. Against multiple baselines, the confidence score head outperforms all baselines and metamodels, automating the largest fraction of the workload at high accuracy targets. Prompted confidence scores can appear well calibrated yet lack the resolution needed for thresholding. Although transferability to unseen datasets is modest, the confidence score head remains the most robust approach. The proposed confidence scoring framework enhances the safety of automating LLM-based agents and can be adapted to gate other properties, including safety or compliance.
\end{abstract}

\section*{Introduction}
AI agents are increasingly deployed in domains such as healthcare~\cite{singhal2023clinical}, finance~\cite{li2023finance}, legal services~\cite{katz2024bar}, and customer support~\cite{brynjolfsson2025work}. Because they process natural language and generate human-like prose, they can assist in tasks such as providing recommendations~\cite{zhao2024recsys}, answering numerical questions over financial reports~\cite{chen2021finqa}, and automating decision-making processes~\cite{wang2024agents}. Moreover, their autonomy extends beyond query-response interactions to multi-step reasoning, planning, and action execution based on dynamic inputs and evolving contexts~\cite{yao2023react,xi2025agents}. Such actions can be as harmless as scheduling a meeting or as high-stakes as denying a loan~\cite{babaei2024credit} or making medical recommendations~\cite{singhal2025expert}.

\begin{figure}[!t]
  \centering
  \includegraphics[width=\linewidth]{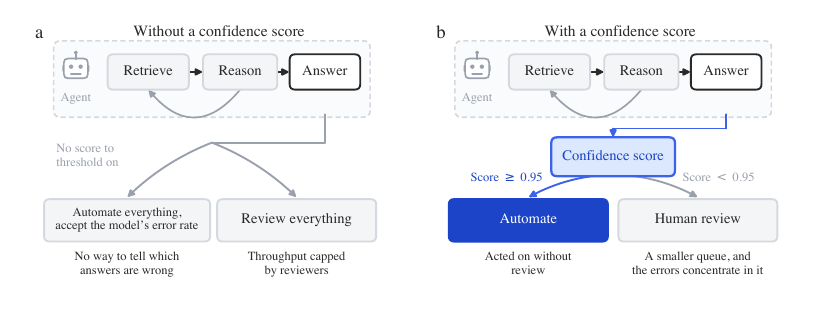}
\caption{\textbf{AI agent decision-making.} The same agent in both panels retrieves, reasons, and produces an answer. \textbf{a)} All answers are acted on, and every wrong answer reaches the action unchecked. Alternatively, every answer can be routed to a reviewer in a human-in-the-loop approach. In this case, throughput is capped by the reviewers, and no answer is acted on without review. \textbf{b)} A confidence score for each answer gates the action. Answers above a threshold are acted on without review, while those below are routed for human review. The threshold value shown is illustrative.}
  \label{fig:workflow}
\end{figure}

Human-in-the-loop review has been the default safeguard, but it is resource-intensive and may not scale with the growing deployment of AI agents. Systems that use agents for decision-making need a scalable way to estimate the probability that each output is correct. Figure~\ref{fig:workflow} illustrates how a confidence scoring framework can be applied to this problem. An ungated agent either acts on every answer, letting its errors reach the action, or sends every answer to a reviewer, capping throughput at human capacity (Fig.~\ref{fig:workflow}a). A confidence score gates the action so that answers scored above a threshold are acted upon and the rest are routed for review (Fig.~\ref{fig:workflow}b).

The estimation of correctness depends on how an agent's output is produced. In AI agents, the output is typically generated by a large language model (LLM)~\cite{xi2025agents}. Most LLMs are transformer decoders~\cite{vaswani2017attention}, called causal models because they generate each token from its conditional probability given the preceding ones~\cite{brown2020fewshot}. Thus, an answer is a sequence of token-level choices, each conditioned on the ones before it. No  quantity produced during generation gives the probability that the answer as a whole is correct.

Many confidence estimation methods aggregate the log-probabilities of the generated tokens into a confidence score for the whole output. A popular approach is to use their average~\cite{jiang2021howcanweknow}. Practitioners also use self-assessment mechanisms, where the model generates an explicit confidence score for its own output~\cite{lin2022teaching,kadavath2022know,tian2023justask}. These scores depend on the model's judgment of its own answer, which can be over-confident and biased. Similarly, LLM-as-Judge approaches~\cite{zheng2023judging}, where a separate model call evaluates the correctness of another call's output, inherit the biases of the judging model. More recently, typed-decision models (introduced as System One models~\cite{laya,jev}), purpose-built for decision-making rather than text generation, replace the generative judging call with a single forward pass that outputs a probability without generating text.

Here, we introduce a supervised approach to estimating confidence scores. We fit two kinds of scorers on correctness labels, which we obtain by grading the model’s answers automatically against reference answers. The first, a confidence score head added to the causal LLM, follows the probing approach~\cite{azaria2023internalstate,liu2024supervised,kossen2024sep}. It is a linear classifier attached to the LLM frozen backbone. The confidence score head reads the final-layer hidden state at the last token and predicts the probability that the generated answer is correct. For proprietary models whose internal states are not exposed, we also introduce metamodels that predict answer correctness based on the LLM’s input and output text. 

We compare both approaches against the token log-probability, a self-score generated with the answer, LLM-as-Judge, and a typed-decision model on SQuAD and TruthfulQA. We use each score as the gate in Fig.~\ref{fig:workflow}b, and set the threshold by the target accuracy the automated answers must reach. We evaluate each score by its discrimination, its calibration, and the fraction of answers it automates at that accuracy target. We extend the comparison to six out-of-domain QA datasets (MRQA 2019 shared task~\cite{fisch2019mrqa}) not used in training. Our main contribution is an operational comparison of probes (such as our confidence-score head), text-only metamodels, and single-pass baselines. Alongside standard discrimination and calibration metrics, we measure the fraction of answers each method automates at a fixed accuracy target, since this target is what sets the threshold of a gate in practice.

\section*{Related work}
Previous studies have investigated when a model's output can be acted on without review in the context of selective prediction, in which a classifier predicts some inputs and abstains on others, so that accepting fewer inputs (i.e., lower coverage) buys higher accuracy~\cite{geifman2017selective}. Learning to defer extends this idea by handing the abstained inputs to a person~\cite{madras2018predict,mozannar2020consistent}. In LLM-based agents, conformal prediction has been used to make the agent ask for help when more than one candidate action remains plausible~\cite{ren2023robots}.

A further line of work, known as risk control, sets the threshold so that the error rate stays below a chosen value, with a guarantee that holds in finite samples. It does so either by extending conformal prediction to monotone losses~\cite{angelopoulos2024crc} or by treating threshold selection as a multiple testing problem~\cite{angelopoulos2021ltt}. Conformal abstention brings this guarantee to LLMs~\cite{abbasiyadkori2024abstention}. The model achieves a target accuracy by selectively abstaining from certain questions, ensuring the error rate on the answers it provides remains strictly below the specified threshold. We do not provide such a guarantee. We select a threshold on a validation sample and report the accuracy the accepted answers reach, with no finite-sample bound on the error rate of the answers a gate automates.

Probing methods extract correctness signals directly from a model's hidden states, using either supervised classifiers~\cite{azaria2023internalstate,liu2024supervised} or unsupervised learning~\cite{burns2023latentknowledge}. Marks and Tegmark show that the truth of a statement is represented linearly in these states and can be recovered using a difference-in-means direction~\cite{marks2024geometry}. Taking a different approach, alternative methods evaluate the semantic consistency of sampled generations directly in the embedding space rather than at the token level~\cite{chen2024inside}. However, this internal signal often concentrates on specific tokens, causing the resulting detectors to transfer poorly across datasets~\cite{orgad2025llmsknow}, a limitation we directly assess in our out-of-domain evaluations. In related work, value heads can predict whether a model knows an answer before generation~\cite{kadavath2022know}, and single-pass probes can approximate more expensive sampling-based measures~\cite{kossen2024sep}. These studies indicate that hidden states encode a correctness signal.

Our confidence score head is a supervised probe of this kind~\cite{liu2024supervised,kossen2024sep}. For closed models, where hidden states are inaccessible, our metamodel adapts the black-box approach of Liu et al.~\cite{liu2024supervised}. It uses sentence embeddings of the text instead of internal states. Prior work demonstrates that training an auxiliary predictor solely on the question and generated answer significantly improves error detection over standard black-box baselines~\cite{ulmer2024apricot}. Our metamodel directly adopts this approach. When benchmarking these estimators, trained probes typically prove most reliable, provided weights and labels are available~\cite{mahaut2024factual}. We confirm this performance ranking and extend the evaluation by measuring these methods at fixed accuracy targets, i.e., the operational thresholds used for gating in practice.

We compare both approaches against several single-pass baselines that require no additional training. First, we extract confidence directly from the generation process as the average token log-probability~\cite{jiang2021howcanweknow}. Next, we evaluate three prompting-based methods: $i)$ verbalized confidence, here called self-score (schema), which prompts the model to state its certainty~\cite{lin2022teaching,tian2023justask}; $ii)$ self-evaluation, which asks the model whether its proposed answer is correct~\cite{kadavath2022know}; and $iii)$ LLM-as-Judge, which assesses the output with a separate model call~\cite{zheng2023judging}. Finally, we include typed-decision models~\cite{laya,jev}, which replace the separate LLM judging call with a single forward pass that outputs a probability. From this category, we evaluate Laya, an open-weight bidirectional encoder, without fine-tuning.

Sampling-based methods estimate confidence by measuring agreement across multiple generations for the same input, either directly~\cite{wang2023selfconsistency} or after semantically grouping answers~\cite{kuhn2023semantic}. However, these methods require repeated decoding for every query, so their computational cost scales with the number of generations. In contrast, the baselines we evaluate and our proposed methods require at most one additional forward pass and no further decoding. Therefore, we exclude sampling-based methods from this study.

Evaluating these confidence scores also requires separating their calibration from their discrimination~\cite{liu2024supervised}. A score is calibrated when its predicted probabilities match the observed frequencies of correctness~\cite{guo2017calibration}. To formalize this distinction, we apply Murphy's decomposition of the Brier score into reliability, resolution (which measures discrimination), and uncertainty~\cite{murphy1973partition}. This distinction matters because miscalibration can be corrected post hoc with a monotonic mapping~\cite{platt1999probabilistic}, whereas a lack of discrimination cannot, since a monotonic mapping leaves the ranking of answers unchanged.

\section*{Results}
\subsection*{Proposed Architectures}

\begin{figure}[!h]
  \centering
  \includegraphics[width=\linewidth]{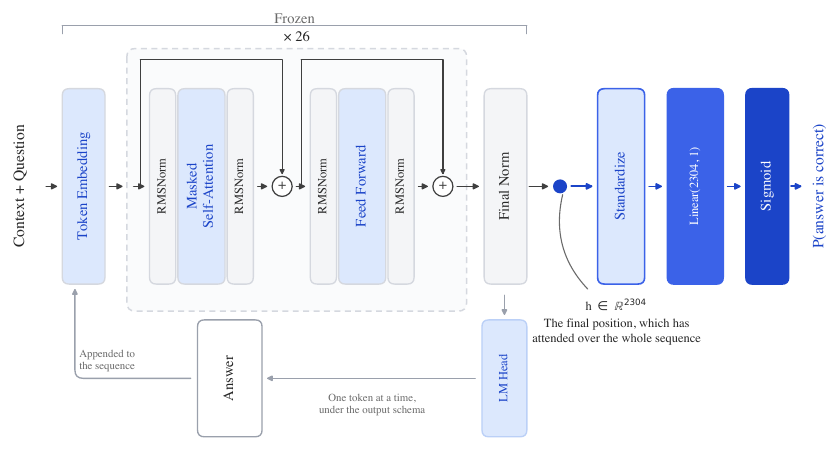}
  \caption{\textbf{LLM architecture and new confidence score head.}
The confidence score head reads the final-layer hidden state $\mathbf{h}\in\mathbb{R}^{2304}$ at the last token of the concatenated prompt and generated answer and maps it to a probability by standardizing it coordinate-wise to $\mathbf{z}$, then applying a linear layer and a sigmoid. The LLM backbone (\texttt{gemma-2-2b-it}, 26 layers, 4-bit) is frozen during training, and only the parameters of the confidence score head are updated. The answer is generated autoregressively under the output schema and appended to the sequence before the confidence score head reads it. The last position attends over the whole sequence so that a single vector can depend on the answer, the question, and the context together.}
  \label{fig:architecture}
\end{figure}

\begin{figure}[H]
  \centering
  \includegraphics[width=3.40in]{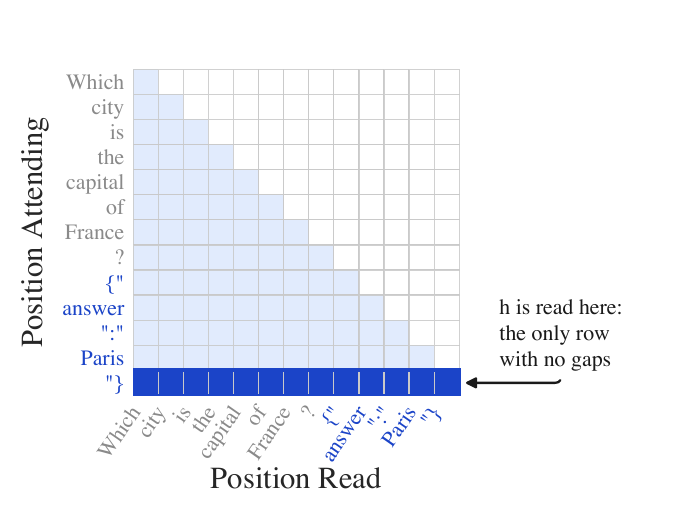}
  \caption{\textbf{The causal attention mask.} Each row shows the positions a token can attend to under causal attention. Only the last token attends to every position in the sequence, so its hidden state $\mathbf{h}$, which the confidence score head reads, can depend on the whole prompt and answer. The example is abbreviated. The generations in this study also contain the \texttt{detailed\_answer} field.}
  \label{fig:mask}
\end{figure}

Figure~\ref{fig:architecture} illustrates the proposed architecture, which pairs a frozen LLM backbone with a lightweight confidence score head of 2{,}305 parameters. During inference, the LLM generates the answer autoregressively. The head reads the final-layer hidden state at the last token of the concatenated prompt and answer to predict a confidence score. Under causal attention, this last position is the only one that attends over the whole sequence (Supplementary Fig.~\ref{fig:mask}). This design reuses the LLM's contextual representations and allows the head to be trained without updating the backbone.

The confidence score $\hat{p}$ is computed as
\begin{equation}
\hat{p} = \sigma(\mathbf{w}^\top \mathbf{z} + b),
\end{equation}
where $\mathbf{w} \in \mathbb{R}^{2304}$ and $b \in \mathbb{R}$ are the learned weights and bias, $\sigma$ is the sigmoid function, and $\mathbf{z}$ is the final-layer hidden state $\mathbf{h}$ standardized coordinate-wise. That is, $z_j = (h_j - \mu_j)/s_j$, where $\mu_j$ and $s_j$ are the mean and standard deviation of coordinate $j$, computed once on the rows the head is fitted on and held fixed thereafter (see Methods).

The confidence score head relies on a model's internal states, which are inaccessible for proprietary models served behind an API. For these models, we introduce a second architecture, the metamodel, which takes the context, question, and answer as text. We embed the three fields separately with a sentence encoder (\texttt{gte-modernbert-base}) and concatenate the embeddings into a single feature vector. To predict answer correctness, we train three classifiers on this vector (logistic regression, a random forest, and XGBoost) and evaluate each as an independent metamodel. Thus, the metamodel requires only the textual inputs and outputs of the API and no access to model weights or log-probabilities (Methods).

\begin{figure}[H]
  \centering
  \includegraphics[width=2.88in]{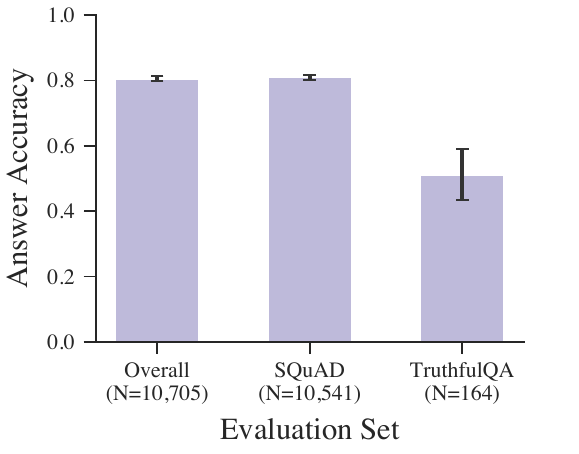}
  \caption{\textbf{LLM answer accuracy on the validation split.} Fraction of answers graded correct, overall and per dataset, with 95\% bootstrap intervals. TruthfulQA contributes 164 of the 10{,}705 answers, so the overall accuracy is close to that of SQuAD.}
  \label{fig:answeracc}
\end{figure}

Both architectures are trained as binary classifiers on labels that indicate whether each generated answer is correct. We evaluate on SQuAD~\cite{rajpurkar2016squad} and TruthfulQA~\cite{lin2022truthfulqa}, both of which provide reference answers that allow every generated answer to be graded automatically. SQuAD is an extractive task large enough to fit the scorers, in which the answer is contained in a given passage. TruthfulQA has no passage, so the model must answer from its own knowledge, choosing among options whose incorrect ones are common misconceptions. We include both so that the scorers are fitted and evaluated on answers of both kinds, rather than on extractive answers alone. For SQuAD, an answer is labeled correct when it contains a reference answer or matches one with a token-level $F1$-score of at least 0.8. For TruthfulQA, it is labeled correct when it selects a correct option (Methods). Under these rules, 81\% of the validation answers are correct (Supplementary Fig.~\ref{fig:answeracc} and Table~\ref{tab:answeracc}).

\subsection*{Classification Performance}
\begin{figure}[!h]
  \centering
  \includegraphics[width=4.04in]{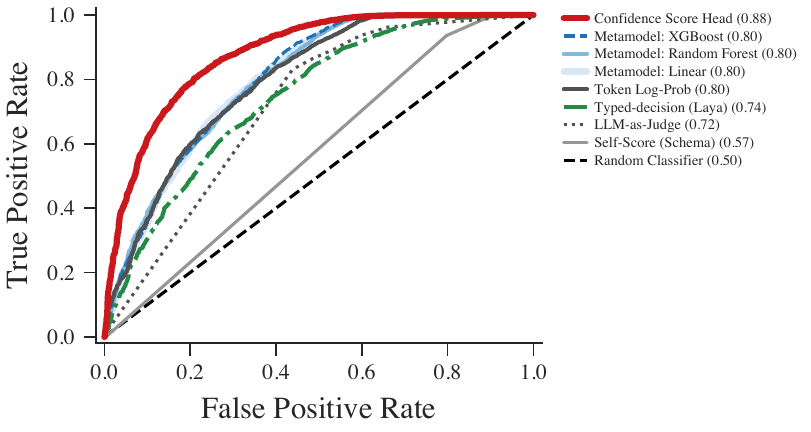}
\caption{\textbf{ROC curves comparing the classification performance of the confidence score estimation approaches.} Values in parentheses indicate each model's ROC-AUC. The confidence score head is shown in red, the three metamodels in blue, the typed-decision model in green, and the remaining baselines in different shades of gray.}
  \label{fig:roc}
\end{figure}

To compare the classification performance of our models, we use ROC-AUC, which measures how well a model distinguishes correct from incorrect answers across all classification thresholds. ROC-AUC provides a single score between 0 and 1, where 1.0 is a perfect model, and 0.5 represents random guessing. In Figure~\ref{fig:roc}, we show ROC curves comparing the classification performance of each confidence-score estimation model.

\begin{figure}[H]
  \centering
  \includegraphics[width=2.75in]{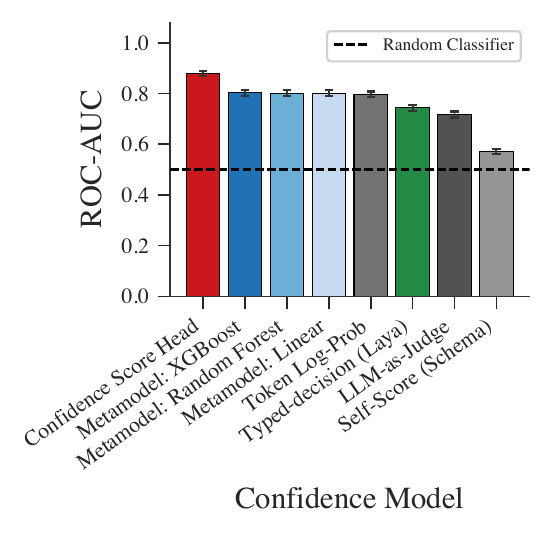}
  \caption{\textbf{ROC-AUC by model.} Error bars are 95\% bootstrap intervals, and the dashed line marks a random classifier. The intervals of the three metamodels overlap.}
  \label{fig:aucbars}
\end{figure}

The confidence score head is the model that best discriminates correct from incorrect LLM answers, followed by the metamodels and the token log-probability. The 95\% bootstrap interval for the ROC-AUC of the confidence score head lies above that of every other model (Supplementary Fig.~\ref{fig:aucbars} and Table~\ref{tab:cutmetrics}). In contrast, the intervals of the metamodels and the token log-probability overlap. Although the metamodels and token log-probability reach the same ROC-AUC (0.80), they require different access to the LLM. The token log-probability requires per-token probabilities generated by the LLM, which a served model behind an API may not return. In contrast, the metamodels need only the input and output text. Thus, the metamodels are the alternative that applies to any LLM, including those whose hidden states are not exposed and that do not return token probabilities.

The token log-probability reaches this performance only when averaged over the tokens of the parsed \texttt{answer} field (Methods). Averaged over the entire generation, which includes schema syntax and the \texttt{detailed\_answer} field, its ROC-AUC falls to 0.64 [0.63, 0.65]. This suggests the score depends on isolating the answer span, which the output schema makes possible by fixing its position in the generation.

The prompting baselines, LLM-as-Judge and the self-score (schema), have the lowest ROC-AUC, with the self-score (schema) approaching random guessing. Both baselines ask the LLM to generate a score for its own answer, whereas the confidence score head maps the final-layer representation directly to a probability using logistic regression. Thus, the gap between the two baselines and the head suggests that the hidden state carries more information about correctness than the LLM can report when prompted to score its own answer, either during generation or in a separate call.

Because the typed-decision model and LLM-as-Judge are both unfitted and read identical inputs, their comparison measures what a purpose-built model adds over prompting the LLM. A paired bootstrap over the validation answers gives a ROC-AUC difference of 0.03 [0.01, 0.04] in favor of the typed-decision model. That is, an encoder built for this decision ranks correctness better than an LLM prompted to grade its own answer. However, it remains below the confidence score head and the three metamodels, which are fitted on the training split of these two datasets. Typed-decision models are designed to answer typed questions without task-specific training~\cite{laya,jev}, so we use Laya~\cite{laya} as released. Fine-tuning it on graded answers could narrow its gap to the fitted models, and we leave this comparison for future work.

\begin{figure}[H]
  \centering
  \includegraphics[width=2.75in]{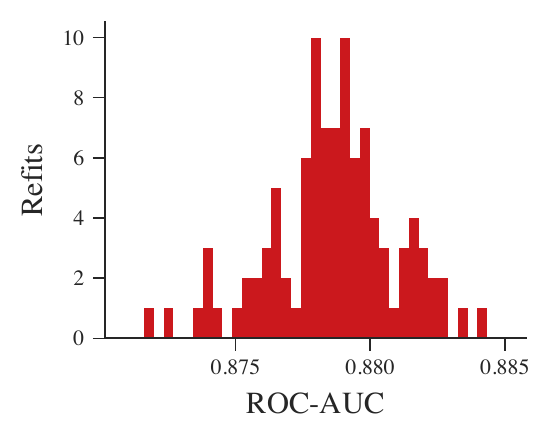}
  \caption{\textbf{ROC-AUC across initialization seeds.} Validation ROC-AUC of the confidence score head over 100 refits that differ only in the random seed (Methods).}
  \label{fig:seedspread}
\end{figure}

We further verify that the ranking of the confidence score head does not depend on its random initialization. Across 100 refits varying only the random seed, its validation ROC-AUC remains between 0.87 and 0.88, within its own [0.87, 0.89] bootstrap interval (Supplementary Fig.~\ref{fig:seedspread}). Thus, the ranking is not an artifact of a particular optimization run.

It is worth noting that the pooled classification metrics are dominated by SQuAD, which contributes 98.5\% of the validation samples. On the TruthfulQA subset alone ($N=164$), the point estimates of the ROC-AUC are 0.66 [0.58, 0.73] for LLM-as-Judge and 0.62 [0.53, 0.70] for the confidence score head (Supplementary Table~\ref{tab:aucperdataset}). Thus, the confidence intervals overlap across most of their range, and the dataset cannot separate the performance of the two methods.

\subsection*{Score Distributions}
\begin{figure}[H]
  \centering
  \includegraphics[width=\linewidth]{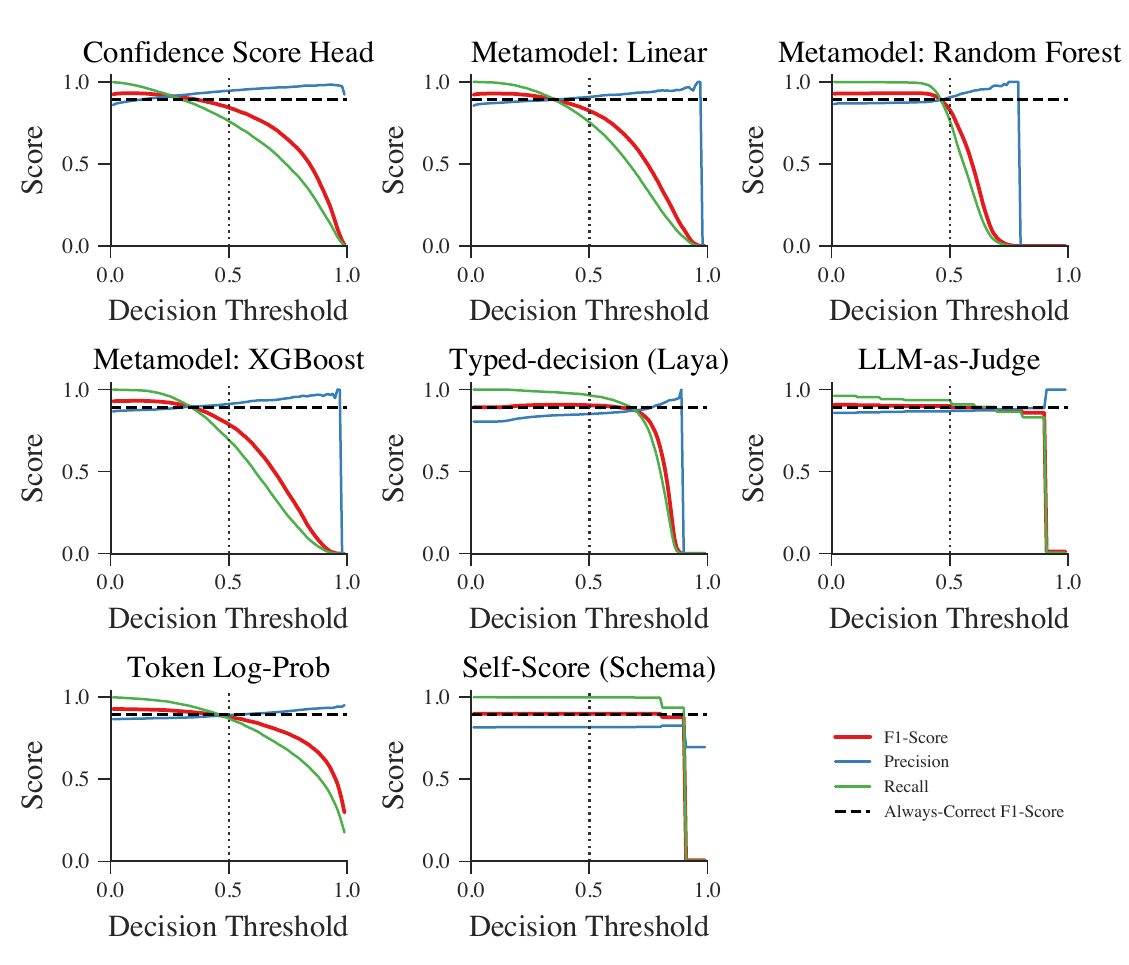}
  \caption{\textbf{Precision, recall, and $F1$-score as a function of the decision threshold.} The positive class is a correct answer. The dashed line marks the $F1$-score of predicting every answer correct, and the dotted line marks the 0.5 threshold. The confidence score head and the metamodels trade precision for recall smoothly, whereas the self-score (schema) and LLM-as-Judge change in steps, because their scores take only a few discrete values.}
  \label{fig:sweep}
\end{figure}

The difference in classification performance is also visible in the score distributions of each method (Supplementary Fig.~\ref{fig:distributions}). A gate needs scores spread over many distinct values, so that a threshold can be placed at any operating point. The confidence score head and metamodels meet this requirement, spreading its scores over a continuous range with the correct and incorrect answers shifted apart. In contrast, the self-score (schema) and LLM-as-Judge suffer from score collapse, i.e., they take only 8 and 13 distinct values across the 10{,}705 validation answers, respectively. Because many answers share the same score, precision and recall change in steps rather than gradually as the threshold varies (Supplementary Fig.~\ref{fig:sweep}), so no threshold on these scores trades coverage for accuracy as a gate would.

\begin{figure}[H]
  \centering
  \includegraphics[width=\linewidth]{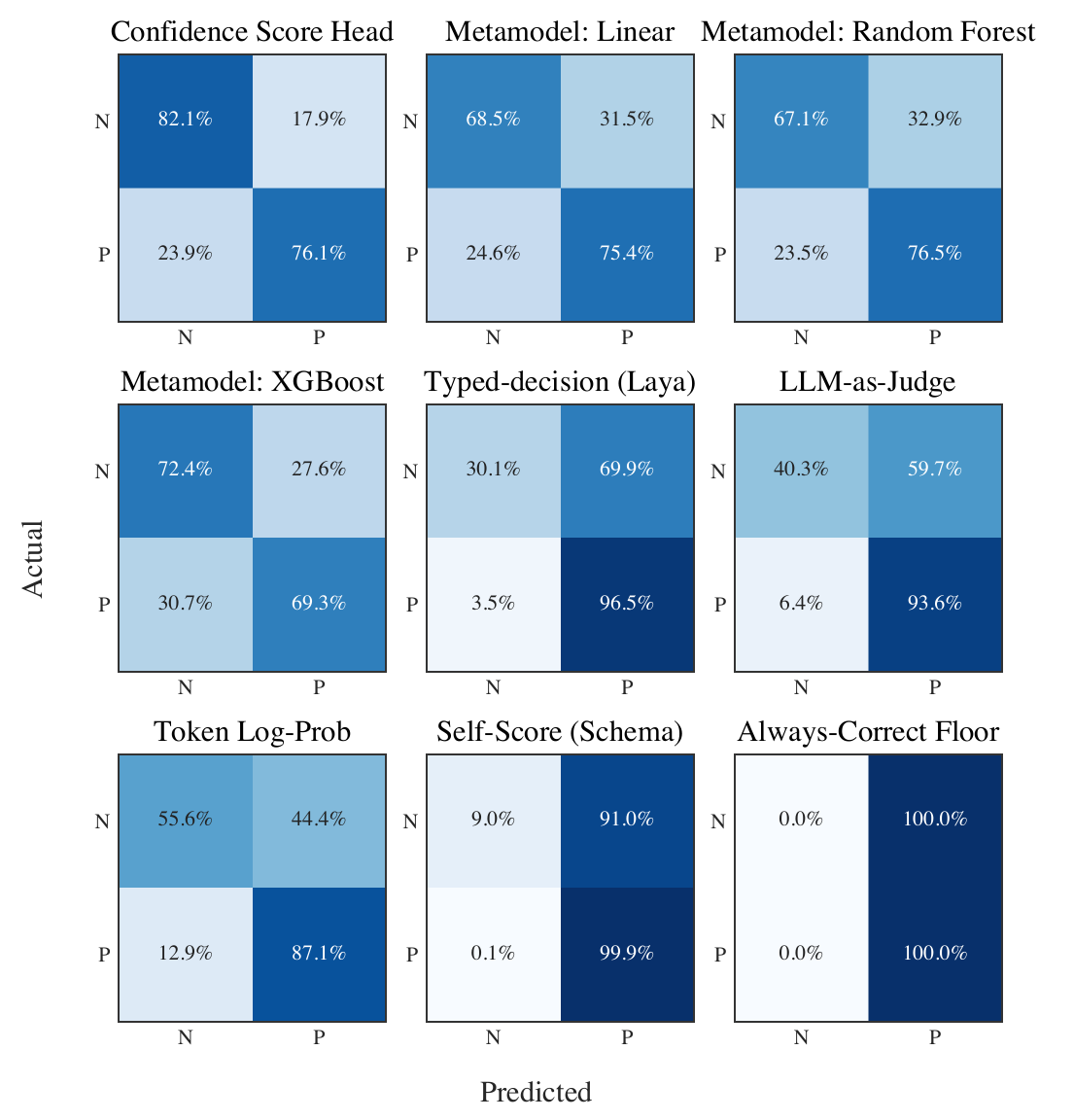}
  \caption{\textbf{Confusion matrices at the 0.5 threshold.} \textbf{P} is an answer graded correct and \textbf{N} an answer graded incorrect. Rows are the actual grade and columns the prediction, and each row sums to 100\%, so the top-right cell is the false positive rate, i.e., the share of incorrect answers labeled correct. The confidence score head rejects 82.1\% of incorrect answers while accepting 76.1\% of correct ones. The self-score (schema), LLM-as-Judge, and typed-decision model label most incorrect answers as correct (91.0\%, 59.7\%, and 69.9\%), whereas the token log-probability labels 44.4\% of them correct while accepting 87.1\% of the correct ones. The last panel shows the always-correct floor.}
  \label{fig:confusionpct}
\end{figure}

\begin{figure}[H]
  \centering
  \includegraphics[width=\linewidth]{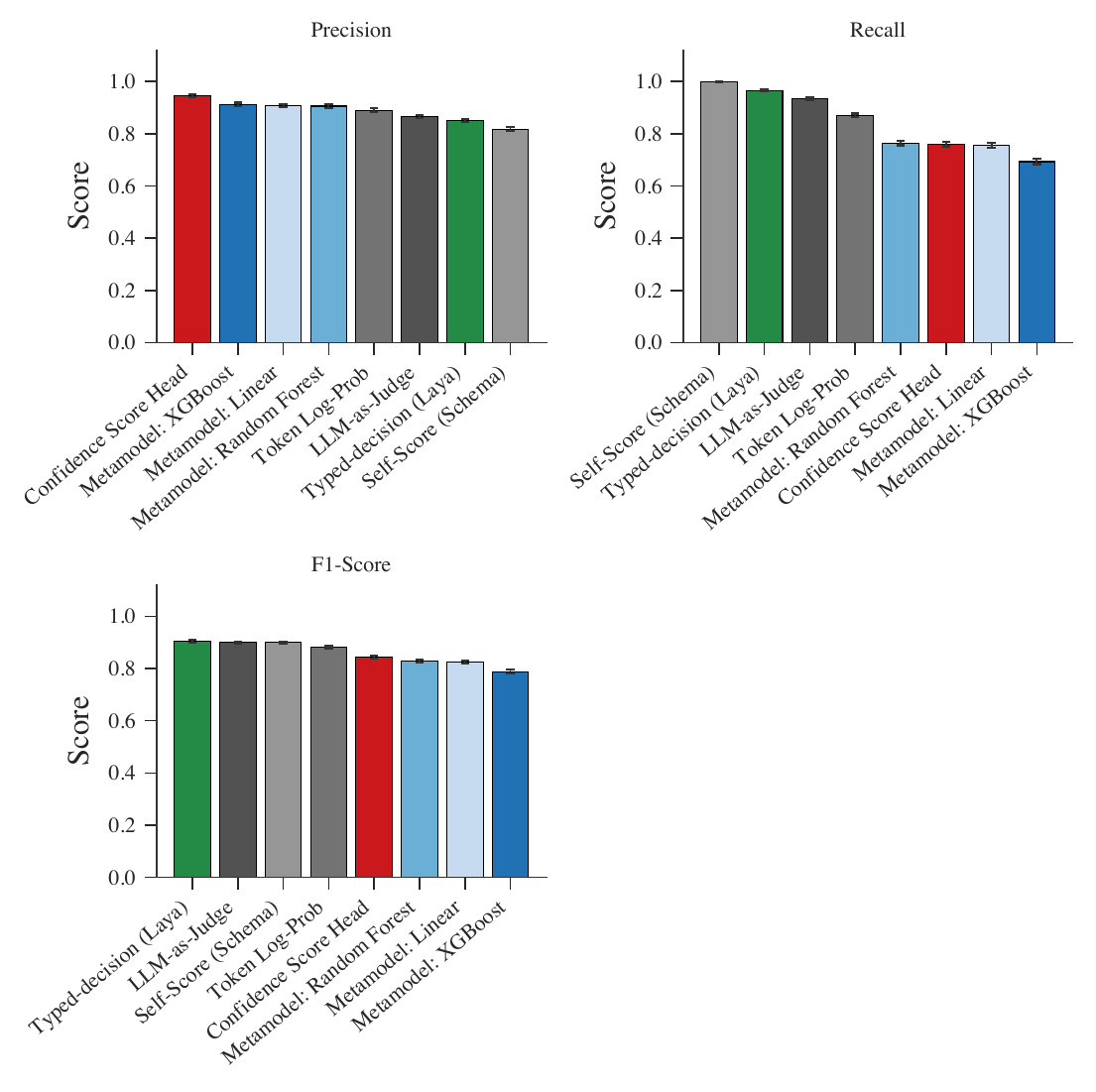}
\caption{\textbf{Precision, recall, and $F1$-score at the 0.5 threshold.} Predicting every answer correct gives a recall of 1.0 and an $F1$-score of 0.89 (the \emph{always-correct} row of Table~\ref{tab:cutmetrics}), so high recall and $F1$-score at this threshold mostly reflect how close a model comes to accepting everything. The self-score (schema), LLM-as-Judge, and typed-decision model are near that floor on both. The confidence score head has the highest precision (0.95) at a recall of 0.76.}
  \label{fig:cutmetrics}
\end{figure}

In a gated system, precision is the accuracy of the automated answers, and recall is the fraction of correct answers that are automated, so recall reaches its maximum when every answer is accepted. At the standard 0.5 threshold, the self-score (schema), LLM-as-Judge, and the typed-decision model label most incorrect answers as correct (Supplementary Figs.~\ref{fig:confusionpct} and~\ref{fig:cutmetrics}, and Table~\ref{tab:cutmetrics}).The self-score (schema) reaches a recall of 1.00 at a precision of 0.82, barely above the 0.80 accuracy of its own answers, so it automates almost every answer, including the incorrect ones. In contrast, the confidence score head holds a precision of 0.95 at a recall of 0.76, filtering out most incorrect answers at the same threshold.

\subsection*{Model Calibration}
A gate acts on a confidence score as if it were a probability of correctness. Calibration, also called reliability, measures whether that reading holds, i.e., whether roughly 60\% of the answers scored 0.6 are correct. A threshold placed on a calibrated score sets a lower bound on the expected accuracy of the automated answers. However, calibration alone does not make a score useful for gating. A model that predicts the base rate for every answer is perfectly calibrated yet carries no information, and a model that concentrates its scores near the base rate approaches this behavior. We therefore measure calibration by the expected calibration error (ECE) and by Murphy's decomposition of the Brier score (Methods), in which the reliability term is the calibration error and the resolution term measures discrimination.

\begin{table}[!h]
  \centering
    \caption{\textbf{Calibration performance of the confidence score estimation approaches.} ECE = expected calibration error. ECE, Brier score, and Reliability are better when lower, whereas Resolution, which measures discrimination, is better when higher. Brackets are 95\% bootstrap intervals. The self-score (schema) is computed on its own generation pass (Methods).}
  \label{tab:calibration}
  \scriptsize
  \setlength{\tabcolsep}{2.6pt}
  \newcommand{\p}{\phantom{0}}
  \begin{tabular}{l r@{\,}l r@{\,}l r@{\,}l r@{\,}l}
    \toprule
    Model & \multicolumn{2}{c}{ECE} & \multicolumn{2}{c}{Reliability} & \multicolumn{2}{c}{Resolution} & \multicolumn{2}{c}{Brier} \\
    \midrule
    Confidence score head & 0.22 & [0.21, 0.22] & 0.07\p & [0.06\p, 0.07\p] & \textbf{0.064} & [0.061, 0.068] & 0.16\p & [0.15\p, 0.16\p] \\
    Metamodel: XGBoost & 0.27 & [0.26, 0.28] & 0.09\p & [0.09\p, 0.10\p] & 0.06\p & [0.05\p, 0.06\p] & 0.191 & [0.188, 0.194] \\
    Metamodel: RF & 0.30 & [0.29, 0.30] & 0.095 & [0.091, 0.099] & 0.06\p & [0.05\p, 0.06\p] & 0.194 & [0.192, 0.196] \\
    Metamodel: linear & 0.23 & [0.23, 0.24] & 0.074 & [0.070, 0.078] & 0.056 & [0.052, 0.059] & 0.175 & [0.171, 0.178] \\
    Token log-prob. & 0.11 & [0.11, 0.12] & 0.029 & [0.026, 0.032] & 0.051 & [0.048, 0.054] & \textbf{0.134} & [0.130, 0.139] \\
    Typed-decision (Laya) & \textbf{0.07} & [0.06, 0.08] & \textbf{0.005} & [0.004, 0.006] & 0.029 & [0.026, 0.032] & \textbf{0.13}\p & [0.13\p, 0.14\p] \\
    LLM-as-Judge & \textbf{0.07} & [0.06, 0.07] & 0.018 & [0.016, 0.021] & 0.029 & [0.026, 0.032] & 0.15\p & [0.14\p, 0.15\p] \\
    Self-score (schema) & \textbf{0.07} & [0.07, 0.08] & 0.006 & [0.005, 0.007] & 0.012 & [0.011, 0.014] & 0.15\p & [0.15\p, 0.16\p] \\
    \midrule
    \textit{constant at base rate} & \textit{0.00} & & \textit{0.00}\p & & \textit{0.00}\p & & \textit{0.16} & \\
    \bottomrule
  \end{tabular}
\end{table}

\begin{figure}[H]
  \centering
  \includegraphics[width=\linewidth]{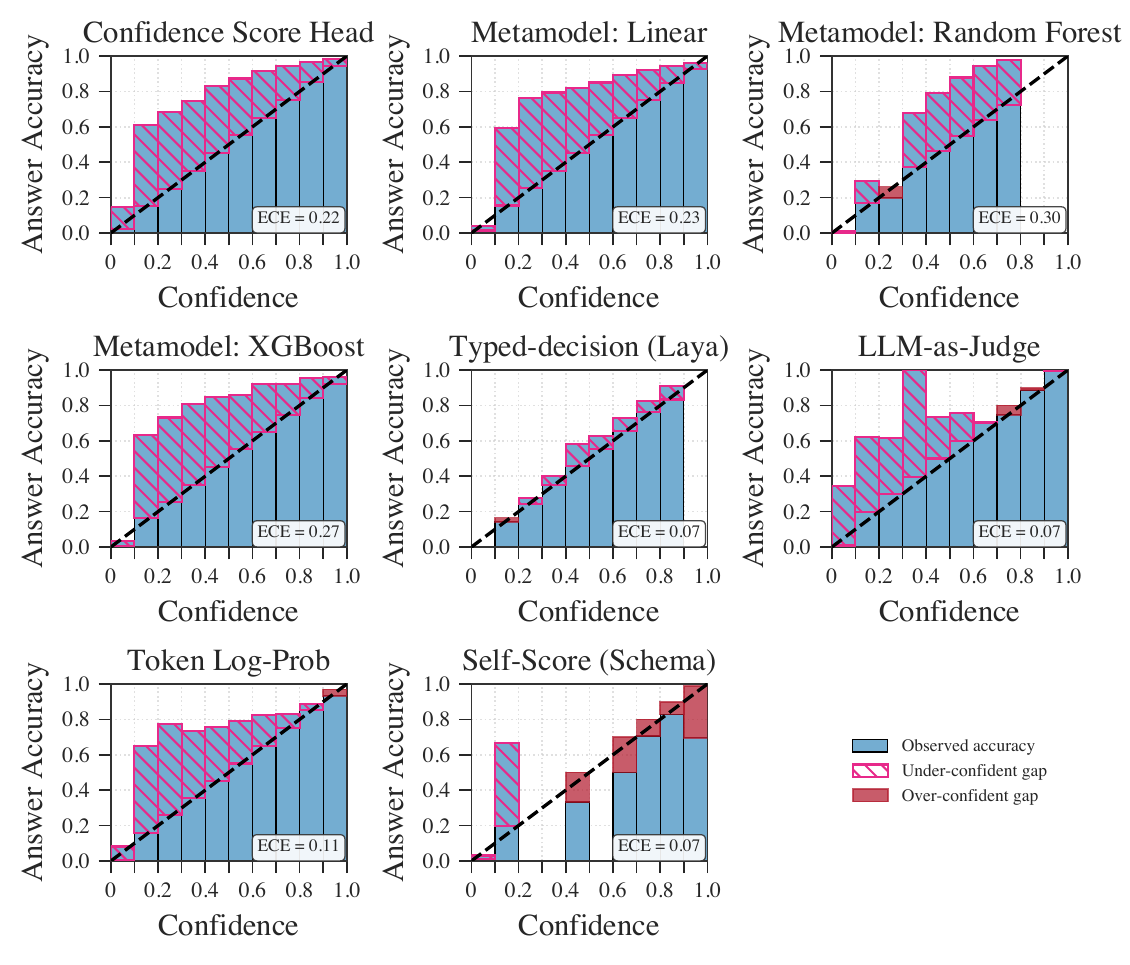}
  \caption{\textbf{Reliability diagrams.} Scores are divided into ten equal-width bins, and each bar shows the answer accuracy within a bin, with the dashed black diagonal marking perfect calibration. In under-confident bins, pink hatching fills the gap from the diagonal up to the accuracy. In over-confident bins, solid red fills the gap from the accuracy up to the diagonal. Each panel reports its ECE, which must be read together with resolution (Table~\ref{tab:calibration}), because predicting the base rate for every answer minimizes it.}
  \label{fig:reliability}
\end{figure}

LLM-as-Judge, the typed-decision model, and the self-score (schema) reach the lowest expected calibration error (ECE), 0.07, with overlapping bootstrap intervals (Table~\ref{tab:calibration}). Murphy's decomposition~\cite{murphy1973partition} explains why a low ECE is not enough. It writes the Brier score as the reliability error minus the resolution plus an uncertainty term fixed by the base rate, so differences between models come almost entirely from the reliability error and the resolution. ECE measures the same gap between score and accuracy as the reliability error, in absolute rather than squared form, and is therefore blind to resolution. The two prompting baselines, LLM-as-Judge and the self-score (schema), keep their reliability errors low (0.018 and 0.006) by clustering their scores near the base rate (Supplementary Fig.~\ref{fig:reliability}), which also keeps their resolution low (0.029 and 0.012). In contrast, the confidence score head has a higher reliability error, 0.07 [0.06, 0.07], but the highest resolution of any method, 0.064 [0.061, 0.068].

\begin{figure}[H]
  \centering
  \includegraphics[width=\linewidth]{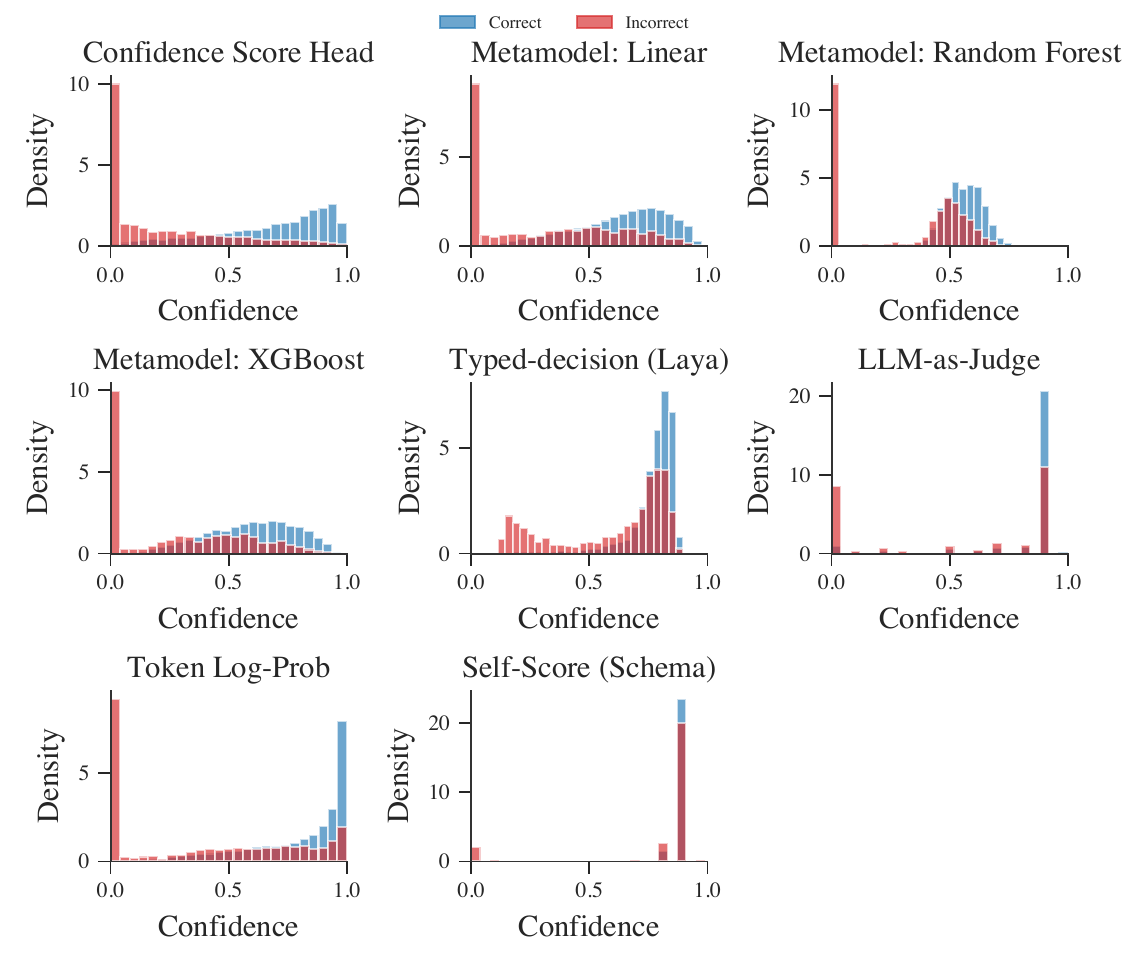}
\caption{\textbf{Score distributions for correct and incorrect answers.} Each model's confidence scores, split by whether the answer was graded correct (blue) or incorrect (red), with each class normalized to its own area. The confidence score head spreads its scores over a continuous range, with the two classes shifted apart. The self-score (schema) and LLM-as-Judge concentrate almost all their mass on a single value and take only 8 and 13 distinct values across the 10{,}705 answers. The typed-decision model also takes a continuous range of values, although its two classes overlap over much of the range.}
  \label{fig:distributions}
\end{figure}

The typed-decision model demonstrates that calibration and resolution can decouple. Its scores span a continuous range (Supplementary Fig.~\ref{fig:distributions}) and its reliability diagram tracks the diagonal across that span, confirming that its low reliability error of 0.005 [0.004, 0.006] is not an artifact of score collapse. Its Brier score of 0.13 [0.13, 0.14] is the lowest of any evaluated method, tied with the token log-probability (paired difference 0.002, 95\% CI [$-0.002$, $0.006$]) and falling below the LLM-as-Judge baseline by a paired margin of 0.01 (95\% CI [0.01, 0.02]). However, its resolution of 0.029 matches that of LLM-as-Judge and is less than half that of the confidence score head. While calibration ensures that scores correspond to empirical frequencies, it does not guarantee the discriminative power required to separate correct from incorrect outputs.

Measured against the Brier score of the constant base-rate predictor (0.16), the models fall into three groups, determined by the balance of their reliability error and resolution. In the first group, the confidence score head matches the constant predictor at 0.16 [0.15, 0.16] because its resolution (0.064) and reliability error (0.07) are nearly equal and cancel out. However, the head ranks answers with a ROC-AUC of 0.88, whereas the constant predictor does not discriminate correct from incorrect answers. The self-score (schema) also matches the constant predictor. In its case, the reliability error (0.006) and resolution (0.012) are both close to zero, so it carries little information about correctness.

In the second group, LLM-as-Judge (0.15), the token log-probability (0.13), and the typed-decision model (0.13) fall below the constant predictor, because their resolution exceeds their small reliability errors. Among them, the token log-probability is the only method that combines a small reliability error with a resolution close to that of the confidence score head. In the third group, the metamodels (0.18 to 0.19) exceed the constant predictor. Their resolution is higher than that of every method except the confidence score head, so their Brier score comes from their large reliability errors, not from a lack of resolution.

The confidence score head and the metamodels are miscalibrated in the same direction, i.e., both are under-confident (Supplementary Fig.~\ref{fig:reliability}). Both are trained with class weights that down-weight the correct answers, and both are fitted on a SQuAD training split whose questions carry a single reference answer, whereas those in the validation split carry several (Methods). As a result, their scores fall below the fraction of answers graded correct at the same score. When a score is read directly as a probability of correctness, under-confidence is the safer error, because borderline answers are routed to human review rather than automated. In contrast, the self-score (schema) is over-confident, i.e., its stated confidence exceeds the observed accuracy at most of the score values it produces. Miscalibration in either direction is also easier to correct than a lack of resolution. Platt scaling~\cite{platt1999probabilistic} reduces the reliability error of the confidence score head from 0.07 to 0.0001 [0.0001, 0.0005] and its ECE from 0.22 to 0.006 [0.005, 0.013] (Supplementary Table~\ref{tab:platt}). Because the rescaling is monotonic, it preserves the ranking of the answers and leaves ROC-AUC unchanged. 

After rescaling, every method has an ECE of at most 0.020, so the methods are nearly equally calibrated and differ mainly in how well they separate correct from incorrect answers. With every reliability error close to zero, the Brier scores are also ordered by resolution. The Brier score of the confidence score head falls to 0.09, the lowest of any method. The metamodels and the token log-probability reach 0.10, whereas the typed-decision model, LLM-as-Judge, and the self-score (schema) remain at 0.13, 0.13, and 0.15, respectively. Thus, the tie between the head and the constant predictor before rescaling is a property of the unscaled scores, not of the information the head carries about correctness.

\subsection*{Automation and Coverage}
\begin{figure}[!h]
  \centering
  \includegraphics[width=\linewidth]{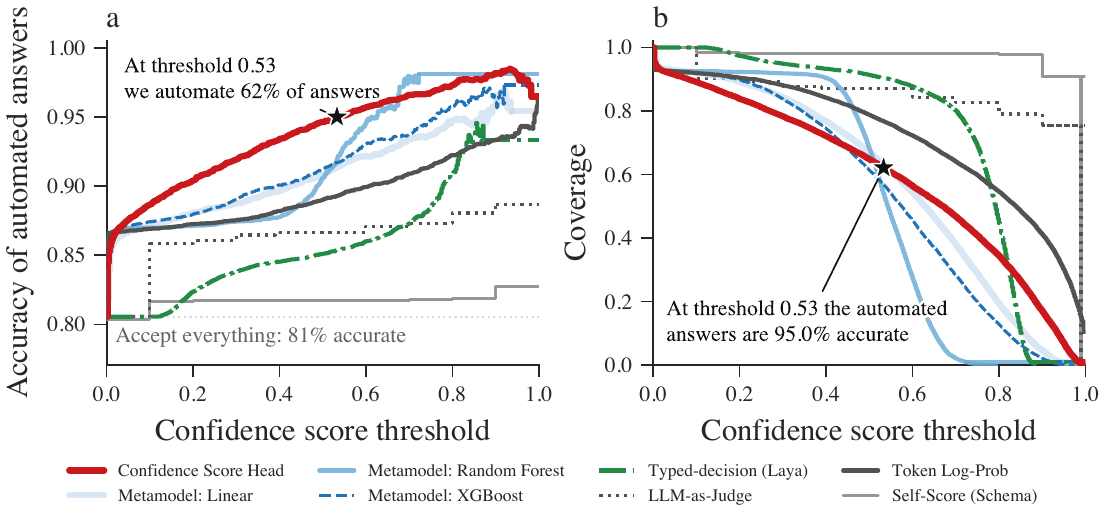}
\caption{\textbf{Automation performance of the confidence score estimation approaches.} \textbf{a)} Accuracy of the automated answers as a function of the acceptance threshold. \textbf{b)} Coverage, i.e., the fraction of answers that are automated, as a function of the acceptance threshold. The star in both panels marks a shared operating point. At threshold $0.53$, the confidence score head's automated answers are 95\% accurate and cover 62\% of the workload.}
  \label{fig:automation}
\end{figure}

Next, we measure how much of the workload each method can automate. In the gated workflow, every answer the agent produces receives a confidence score, and a single threshold splits the answers into two streams. Answers scored at or above the threshold are acted on without review, and those below it are routed to a person (Fig.~\ref{fig:workflow}b). A higher threshold generally makes the accepted answers more accurate, but fewer answers are accepted. Each method has its own trade-offs between the accuracy of the accepted answers and the coverage, i.e., the fraction of the workload accepted. Figure~\ref{fig:automation} and Table~\ref{tab:automation} show this trade-off for each method.

The confidence score head automates the largest fraction of the workload at high accuracy targets, and its lead grows as the target increases(Table~\ref{tab:automation}). At the 90\% target, the metamodels, the token log-probability, and the typed-decision model still automate more than half of the answers. However, their coverage falls sharply at 95\%, where the head keeps 62.1\% of the workload, compared with at most 22.7\% for any other method. This drop is not visible in ROC-AUC, which averages the ranking over all thresholds, whereas a high accuracy target depends only on the highest-scored answers. The other methods still mix incorrect answers into their top scores, so reaching 95\% requires a high threshold that accepts few answers. The prompting baselines automate almost nothing at high accuracy targets. LLM-as-Judge accepts the same 0.5\% at all three targets, because its few distinct score values allow only a few thresholds, and the self-score (schema) accepts none.

\begin{table}[h]
  \centering
  \caption{\textbf{Automation performance at different accuracy targets.} Each cell is the coverage, i.e., the percentage of answers accepted at the threshold that reaches the given accuracy target. Accepting every answer gives 81\% accuracy at 100\% coverage.}
  \label{tab:automation}
  \small
  \begin{tabular}{lrrr}
    \toprule
    Model & $\geq90\%$ & $\geq95\%$ & $\geq97\%$ \\
    \midrule
    Confidence score head & \textbf{85.5\%} & \textbf{62.1\%} & \textbf{38.2\%} \\
    Metamodel: XGBoost          & 75.2\% & 17.8\% & 4.2\% \\
    Metamodel: RF               & 73.1\% & 22.7\% & 4.6\% \\
    Metamodel: linear           & 74.6\% & 9.0\%  & 2.8\% \\
    Token log-prob.             & 71.8\% & 15.4\% & 8.4\% \\
    Typed-decision (Laya)       & 57.3\% & 0.2\%  & 0.1\% \\
    LLM-as-Judge                & 0.5\%  & 0.5\%  & 0.5\% \\
    Self-score (schema)         & 0.0\%  & 0.0\%  & 0.0\% \\
    \bottomrule
  \end{tabular}
\end{table}

The severity of a mistake can set the accuracy target. When an error is costly to reverse, such as an incorrect entry in a patient record, a higher target keeps errors among the automated answers rare, at the cost of sending more answers to review. When an error is cheap to catch or undo, such as a mislabeled support ticket, a lower target automates more of the workload. At the 95\% target, the confidence score head automates 62.1\% of the workload, and the errors among those answers amount to about 3\% of all answers. The remaining errors end up in the review queue, where 43\% of answers are incorrect, compared with 19\% across the workload as a whole. At the same target, the best metamodel (random forest) automates 22.7\% of the workload and leaves a review queue in which 24\% of the answers are incorrect. Thus, the confidence score head concentrates human review on the answers most likely to be incorrect.

The thresholds in Table~\ref{tab:automation} are selected and evaluated on the same validation answers, so they can fit the particularities of that sample and miss their accuracy target on answers not seen before. To test whether the reported coverages are optimistic for this reason, we divide the validation split into ten folds, select each threshold on nine of them, and evaluate it on the remaining one (Methods). The coverages of the confidence score head remain almost unchanged (Supplementary Table~\ref{tab:heldout}), and its automated answers reach every accuracy target out of sample. Thus, its coverages in Table~\ref{tab:automation} are not materially optimistic. In contrast, the linear metamodel, with its threshold selected for 97\% accuracy, reaches 94.2\% on the held-out folds, so the error rate among its automated answers is nearly twice the 3\% it was set for. However, this test assumes that the answers seen in production come from the same distribution as the validation split, and we relax this assumption in the out-of-domain analysis below.

\subsection*{Generalization to unseen datasets}
All previous results were measured on held-out questions from the datasets on which the fitted models were trained. Next, we apply the confidence score head and the metamodels without refitting, along with the unfitted baselines, to six QA datasets from the MRQA 2019 shared task~\cite{fisch2019mrqa} not used in training (Methods). On these datasets, the backbone answers 65\% of the questions correctly, against 81\% on SQuAD and TruthfulQA (Supplementary Tables~\ref{tab:transferbaserate} and~\ref{tab:answeracc}), so every score is evaluated against a lower base rate.

The confidence score head keeps its ranking advantage out-of-domain. It reaches the highest macro ROC-AUC of the eight models (Table~\ref{tab:transfer}) and the highest ROC-AUC on each of the six datasets (Supplementary Table~\ref{tab:transferrocauc}). Its paired margin in macro ROC-AUC over the token log-probability, the strongest baseline by pooled ROC-AUC, is 0.09 [0.07, 0.10], against 0.08 between the same two models in domain, and its paired margin over every other model excludes zero. Thus, the signal the confidence score head reads from the backbone's hidden state carries over to datasets not used in training.

\begin{table}[h]
  \centering
  \caption{\textbf{Discrimination on six datasets not used in training.} Macro ROC-AUC is the unweighted mean over the six datasets, and pooled ROC-AUC is computed over all 8{,}724 answers at once. Brackets are 95\% bootstrap intervals, drawn on a shared resample so that differences between models are paired. The in-domain column repeats the ROC-AUC of Fig.~\ref{fig:roc}, with intervals from Supplementary Table~\ref{tab:cutmetrics}.}
  \label{tab:transfer}
  \small
  \setlength{\tabcolsep}{4pt}
  \begin{tabular}{l r@{\,}l r@{\,}l r@{\,}l}
    \toprule
    Model & \multicolumn{2}{c}{Macro} & \multicolumn{2}{c}{Pooled} & \multicolumn{2}{c}{In domain} \\
    \midrule
    Confidence score head & \textbf{0.82} & [0.81, 0.83] & \textbf{0.83} & [0.83, 0.84] & \textbf{0.88} & [0.87, 0.89] \\
    Metamodel: XGBoost    & 0.73 & [0.72, 0.74] & 0.74 & [0.73, 0.75] & 0.80 & [0.79, 0.81] \\
    Metamodel: RF         & 0.73 & [0.72, 0.74] & 0.74 & [0.73, 0.75] & 0.80 & [0.79, 0.81] \\
    Metamodel: linear     & 0.73 & [0.72, 0.74] & 0.74 & [0.73, 0.76] & 0.80 & [0.79, 0.81] \\
    Token log-prob.       & 0.73 & [0.72, 0.75] & 0.76 & [0.75, 0.77] & 0.80 & [0.79, 0.81] \\
    Typed-decision (Laya) & 0.68 & [0.66, 0.69] & 0.70 & [0.69, 0.72] & 0.74 & [0.73, 0.76] \\
    LLM-as-Judge          & 0.69 & [0.68, 0.70] & 0.71 & [0.70, 0.72] & 0.72 & [0.71, 0.73] \\
    Self-score (schema)   & 0.60 & [0.59, 0.61] & 0.61 & [0.60, 0.61] & 0.57 & [0.56, 0.58] \\
    \bottomrule
  \end{tabular}
\end{table}

The threshold, however, transfers less well than the ranking. With the threshold held at 0.5, the confidence score head reaches a precision of 0.88 [0.87, 0.89] out-of-domain, against 0.95 [0.94, 0.95] in domain (Supplementary Tables~\ref{tab:transferprecision} and~\ref{tab:cutmetrics}). To separate the threshold from the ranking, we also select, on the out-of-domain answers themselves, the lowest threshold whose accepted answers reach each accuracy target (Table~\ref{tab:transferautomation}). Because each threshold is selected on the answers it is evaluated on, this is the best case for a re-selected threshold. Even in this best case, the head automates 20.7\% of the answers at the 95\% target, about a third of its in-domain 62.1\%.

\begin{table}[h]
  \centering
  \caption{\textbf{Automation performance at different accuracy targets on six datasets not used in training.} Each cell is the coverage, i.e., the percentage of answers accepted at the lowest threshold whose automated answers reach the given accuracy target. Accepting every answer gives 65\% accuracy at 100\% coverage. As in Table~\ref{tab:automation}, each threshold is selected on the same answers it is evaluated on. The in-domain values are given in Table~\ref{tab:automation}.}
  \label{tab:transferautomation}
  \small
  \begin{tabular}{lrrr}
    \toprule
    Model & $\geq90\%$ & $\geq95\%$ & $\geq97\%$ \\
    \midrule
    Confidence score head & \textbf{39.1\%} & \textbf{20.7\%} & \textbf{9.0\%} \\
    Metamodel: XGBoost    & 5.5\%  & 0.4\% & 0.4\% \\
    Metamodel: RF         & 0.3\%  & 0.3\% & 0.0\% \\
    Metamodel: linear     & 2.6\%  & 0.4\% & 0.3\% \\
    Token log-prob.       & 1.5\%  & 0.3\% & 0.1\% \\
    Typed-decision (Laya) & 10.5\% & 0.0\% & 0.0\% \\
    LLM-as-Judge          & 0.4\%  & 0.0\% & 0.0\% \\
    Self-score (schema)   & 0.0\%  & 0.0\% & 0.0\% \\
    \bottomrule
  \end{tabular}
\end{table}

Two changes contribute to this drop. First, the LLM answers fewer questions correctly with an accuracy of 65\% out-of-domain compared to 81\% in-domain. The accuracy of the answers accepted above a threshold depends on this base rate, so the same threshold yields a lower precision. Holding the gate to the same accuracy target also generally requires a stricter threshold, which lowers the coverage of every method. Second, the score distributions shifted. The recall of the confidence score head at 0.5 fell from 0.76 to 0.63, and its pooled ROC-AUC fell from 0.88 to 0.83. Our data do not separate the two effects. Even when the threshold is selected on the new datasets, the confidence score head automates 20.7\% of the answers at the 95\% target, against at most 0.4\% for the other methods (Table~\ref{tab:transferautomation}).

\section*{Discussion}
Our results show that an LLM's hidden states carry a more discriminative signal of correctness than the model can verbalize. A lightweight confidence score head reading those states ranks correct answers above incorrect ones well enough to automate most of the workload at a 95\% accuracy target on the datasets it was fitted on, and about a fifth of it on datasets not used in training. This aligns with the probing literature suggesting that LLMs internally encode truthfulness~\cite{azaria2023internalstate,burns2023latentknowledge}, but challenges the reliance on LLM-as-Judge and on verbalized scores such as the self-score (schema).

Access to those states, however, is not a precondition for computing confidence scores. The metamodels read only the text going in and out of the model, with no internal state or log-probability, and still recover about half of the confidence score head's advantage over LLM-as-Judge. Thus, part of the signal the prompting baselines miss is recoverable from outside a proprietary model. On the other hand, the metamodels reach the same ROC-AUC as the token log-probability of the answer.They are therefore preferable when a served model returns neither per-token probabilities nor internal states. Even then, their gap to the head is large in automation. At a 95\% accuracy target, the metamodels automate 9--23\% of the workload, against 62\% for the confidence score head.

The typed-decision model sits between the prompting baselines and our methods. Its ROC-AUC is higher than that of the prompted LLM-as-Judge but lower than those of the confidence score head, the metamodels, and the token log-probability. Thus, replacing the generative judging call with a purpose-built typed decision recovers part of the missing signal, though not all of it. Unlike the prompting baselines, its calibration does not come from scores collapsing onto a few values. However, its resolution remains weak. A model can therefore be well calibrated and still lack the discriminative power that gating requires, which is why calibration and resolution must be evaluated together.

The prompting baselines also reach a low ECE. The self-score (schema) and LLM-as-Judge methods have the lowest ECE, driven by predictions that cluster near the base rate instead of separating correct from incorrect answers. Additionally, their scores take only a few discrete values. Our confidence score head and metamodels, in contrast, have ECEs three to four times higher, but they show a much higher resolution. Their miscalibration is under-confidence. When their score is read as a probability, borderline answers go to review instead of being acted on. Platt scaling after training corrects this calibration error. However, no monotonic rescaling can improve how a score ranks the answers.

The confidence score head and the metamodels are fitted on correctness labels from the training split. In contrast, the token log-probability, the self-score (schema), LLM-as-Judge, and the typed-decision model operate zero-shot. All models are scored on the held-out validation split. The metamodels give the cleanest measure of what this supervised training can offer. They read the same text as the prompting baselines but produce better scores because they were fitted. However, the advantage is smaller out-of-domain. On six QA datasets not used in training, the metamodels reach a macro ROC-AUC of 0.73, against 0.69 for LLM-as-Judge and 0.68 for the typed-decision model. Still, by about half their in-domain margin, and they no longer automate a meaningful fraction of the workload at the 95\% target. The confidence score head maintains an advantage over the token log-probability of the size it has in-domain, and its paired margin over every other model is non-zero.

Our study has several limitations. All experiments use a single frozen backbone, \texttt{gemma-2-2b-it} at 4-bit precision, and a more accurate model would generate fewer incorrect answers to train on. LLM-as-Judge and the self-score are produced by that same 2B model, so they bound what this backbone can say about its own answers rather than what a larger or purpose-built judge would reach. TruthfulQA contributes only 1.5\% of the validation answers, so the pooled results reflect extractive QA on SQuAD. On TruthfulQA alone, the 95\% intervals of the methods' ROC-AUC overlap over most of their range, so this dataset is too small to rank them (Supplementary Table~\ref{tab:aucperdataset}). Transfer is measured within question answering, not across task formats or models. Both datasets are public and may appear in the backbone's pretraining data, which would affect every score we compare. Finally, we evaluate only a linear head and leave other architectures for future work.

Deploying machine learning models to aid decision-making in real-world applications may introduce systemic risks. For example, model performance can degrade silently under distribution shift, and it can be uneven across demographic subgroups (e.g., gender or race). Moreover, the accuracy targets we evaluate are population-level averages, which do not guarantee that an individual answer will meet them. Thus, avoiding silent failures when deploying a confidence-score model requires practices such as monitoring data drift and continuous retraining. A fraction of high-confidence answers should also be reviewed by humans, both as a proxy for drift and as data for retraining. Calibration may need the same monitoring, followed by recalibration. Our out-of-domain results illustrate this need. The 0.5 threshold no longer reaches 95\% accuracy on datasets not used in training. In contrast, a threshold selected again on those datasets does, at about a third of the in-domain coverage (Table~\ref{tab:transferautomation}).

Beyond factual correctness, the same framework could gate other properties of model outputs. Replacing the correctness labels with safety or compliance annotations would turn the confidence score head or the metamodel into a risk score that an agent could check before executing an action. As AI agents are deployed in high-stakes domains~\cite{singhal2023clinical,katz2024bar,li2023finance}, asking a model to assess its own outputs is not enough to decide which of them to act on. Our results show that a score read from the model's internal states, or learned from its inputs and outputs, provides the resolution that such decisions require.
\section*{Methods}

\subsection*{Datasets and splits}
Questions come from SQuAD v1.1~\cite{rajpurkar2016squad}, which pairs an extractive question with a passage containing the answer, and TruthfulQA~\cite{lin2022truthfulqa}, which pairs a question with correct and incorrect options written to elicit common misconceptions. For SQuAD, we keep the dataset's own train/validation split, which shares no articles between the two sides. We key each question by a hash of its text and its passage, which merges the 166 training and 29 validation rows that repeat an identical pair under different identifiers, leaving 87{,}433 and 10{,}541 questions. TruthfulQA ships a single split, from which we hold out 20\% of its 817 questions (seed 42), stratified by its 38 categories. A 15\% development slice of the training rows, stratified by correctness, is used for early stopping of the confidence score head and for hyperparameter selection of the metamodels (Table~\ref{tab:dataset_splits}).

\begin{table}[!h]
  \centering
  \caption{\textbf{Sample sizes for each dataset split.} The development slice is drawn from the pooled training rows, stratified by correctness, and subdivides the training total.}
  \label{tab:dataset_splits}
  \begin{tabular}{lcc}
    \toprule
    Dataset & Training & Validation \\
    \midrule
    SQuAD & 87{,}433 & 10{,}541 \\
    TruthfulQA & 653 & 164 \\
    \midrule
    Total & 88{,}086 & 10{,}705 \\
    \quad trained on & 74{,}873 & \\
    \quad development & 13{,}213 & \\
    \bottomrule
  \end{tabular}
\end{table}

SQuAD questions are presented as the passage followed by the question. TruthfulQA questions are presented as the list of options, one per line, above the question, in a deterministic shuffle seeded by the question text and identical across the answering, self-score, and judgment passes. Example prompts are given in the Supplementary Material.

\subsection*{Out-of-domain evaluation}
We evaluate the frozen models on six out-of-domain subsets of the MRQA 2019 shared task~\cite{fisch2019mrqa}, namely BioASQ, DROP, DuoRC, RACE, RelationExtraction, and TextbookQA. We use only the test split of each dataset (8{,}724 questions in total) and no MRQA training data. Because MRQA records include a context passage, the SQuAD prompt template and grading rules apply without modification. No prompt is truncated in any of the six datasets.

The confidence score head, its 0.5 threshold, and the three metamodels are applied without refitting, with the same backbone and grader used in the domain, and the token log-probability is computed over the answer span as before. We report the macro ROC-AUC, i.e., the unweighted mean over the six datasets, because RelationExtraction alone contributes about a quarter of the pooled answers. Bootstrap intervals resample answers within each dataset, with all models evaluated on the same resample so that differences between models are paired.

\subsection*{Model and generation}
The backbone is \texttt{gemma-2-2b-it}~\cite{gemma2}, an open-weight, instruction-tuned decoder-only model with 26 layers and a hidden size of 2{,}304, loaded in 4-bit NF4 with double quantization~\cite{dettmers2023qlora}, using transformers 5.9.0 and bitsandbytes 0.46.1. The model is not fine-tuned, and the same frozen weights produce the answers, hidden states, token log-probabilities, self-score (schema) outputs, and judgments. Answers are generated greedily with at most 128 new tokens.

\subsection*{Compute}
All generation with the \texttt{gemma-2-2b-it} backbone runs on a single NVIDIA A100-SXM4 (40\,GB) GPU, while the encoders and the fitted models run on commodity CPU hardware. The extraction sweeps over the 88{,}086 training and 10{,}705 validation questions take 6 GPU-hours in total, and fitting the confidence score head on the cached features takes under two minutes.

\subsection*{Constrained decoding}
Every generation in this study is produced under a JSON schema enforced at decode time~\cite{willard2023outlines}. Enforcement constrains which tokens can be emitted but does not force the generation to complete the schema, and 1.9\% of answering-pass generations stop while a string is still open (see Grading).

Answers are generated under a schema with fields \texttt{answer} and \texttt{detailed\_answer}, both constrained to be non-empty strings, and only \texttt{answer} is graded. The answering schema is given below. 

\begin{lstlisting}[style=promptstyle]
{
  "answer": <string>,
  "detailed_answer": <string>
}
\end{lstlisting}

\subsection*{Grading}
The graded answer is the value of the \texttt{answer} field after we remove surrounding quotation marks and any leading separators, decoding artifacts that the LLM introduces under the enforced schema. The grader, i.e., the rules described below, assigns each graded answer a single score between 0 and 1 and labels it correct if that similarity score is at least 0.8. We evaluate every model in our comparison against these labels, with the self-score (schema) graded on its own generation pass. The metamodels, LLM-as-Judge, and the typed-decision model receive this same text as the answer, alongside the context and question described in their respective subsections.

For SQuAD, training questions have a single reference answer, and validation questions have an average of 1.7 distinct references from different annotators. We compare the LLM answer with each reference and take the highest score. A comparison scores 1 when the reference appears in the answer as a contiguous run of tokens and the answer is no longer than five times the reference, or the reference plus fifteen tokens, whichever is larger, so that an answer echoing the entire passage is not scored as correct. Otherwise, the answer is scored by token-level $F1$-score against that reference.

For TruthfulQA, the answer is matched to its nearest listed option by token-level $F1$-score and scores 1 when that option is one of the correct references and 0 otherwise. Questions list 7.0 options on average, 3.1 of them correct, so a random choice scores about 0.45.

Containment accepts an answer that states the reference and adds words around it, which exact match would reject, and the threshold of 0.8 on the token-level $F1$-score labels partially correct answers as incorrect. The grader is therefore a relaxed form of exact match that still restricts how the model may phrase an answer. Both rules apply equally to every score we compare, since the same grader produces the labels for all of them.

Generations that do not parse are not discarded. We close the open string and brace and grade the recovered \texttt{answer} field like any other, so an empty field is graded incorrect. Most of these generations stop early. Only six validation and 63 training generations reached the 128-token limit, all but two in a repetition loop.

The resulting base rate, the fraction of answers labeled correct, is 0.75 on the training split and 0.81 on the validation split (Supplementary Table~\ref{tab:answeracc}). Graded against a single reference, as the training questions are, the validation base rate falls to 0.74, close to the training split's 0.75, and the confidence score head's reliability error falls by about half, to 0.03.

\subsection*{Training the confidence score head}
The confidence score head reads the final-layer hidden state of the backbone at the final token of the prompt-plus-generation sequence, taken from the same teacher-forced forward pass that yields the token log-probabilities. Hidden states are standardized coordinate-wise with means and standard deviations computed on the rows the head is fitted on, excluding the held-out development slice. The head is a linear map from the standardized hidden state to a logit, i.e., logistic regression on the backbone's representation, chosen so that its predictive power can be attributed to the representation rather than to the classifier, and is trained with the class-weighted binary cross-entropy
\begin{equation}
\mathcal{L} = - w \cdot y \cdot \log(\hat{p}) - (1-y) \cdot \log(1-\hat{p}),
\end{equation}
where $y$ is the binary correctness label, $\hat{p}$ is the predicted probability, and $w$ is the weight for the positive class, set to the ratio of negative to positive examples in the training data. Optimization uses AdamW with a learning rate of $10^{-3}$, weight decay of $10^{-2}$, and a batch size of 64, for up to 200 epochs with early stopping on the development loss (patience 20).  Training stopped at epoch 30, and we kept the weights from epoch 10, when the development loss reached its minimum.

We also refit the head 100 times with different random seeds, holding the features, splits, epoch budget, early-stopping rule, and architecture fixed.

\subsection*{Metamodels}
Each metamodel predicts whether the LLM's answer is correct from the text of the context, question, and answer alone, with no access to the LLM's weights or internal states. Each field is embedded independently with \texttt{gte-modernbert-base}~\cite{li2023gte}, built on ModernBERT~\cite{warner2024modernbert}, taking the final-layer hidden state at the CLS token as the embedding. For TruthfulQA, which has no passage, the listed options fill the context field. The longest field is 894 tokens, within the encoder's 8{,}192-token limit, so no input is truncated. The feature vector is the concatenation of the three embeddings,
\begin{equation}
\mathbf{f} = [\mathbf{e}_{\text{context}}; \mathbf{e}_{\text{question}}; \mathbf{e}_{\text{answer}}] \in \mathbb{R}^{2304}.
\end{equation}

On $\mathbf{f}$, we separately train three classifiers (logistic regression, a random forest, and XGBoost), and the predicted probability of each serves as its confidence score. All three are fitted on the training split, with hyperparameters selected on the development slice, and evaluated on the validation split. No early stopping is used.

All three correct for the 3:1 imbalance between correct and incorrect answers. Logistic regression weights the classes by inverse frequency. For the random forest and XGBoost, the correction is selected by grid search on the development slice. For the random forest, the grid is over the weight on the incorrect class paired with the minimum leaf size, $\{(\text{balanced},2),\,(2,12),\,(3,20),\,(5,20),\,(7,20),\,(10,20)\}$, and for XGBoost over \texttt{scale\_pos\_weight}, the weight on the correct class, $\{0.334,\,0.2,\,0.1,\,0.05,\,0.02,\,0.01\}$, where 0.334 is the inverse class frequency of the training split. We select the candidate that minimizes $|\mathrm{TPR}-\mathrm{TNR}|$ on the development slice at a threshold of 0.5. The selected settings are \texttt{scale\_pos\_weight}~$=0.1$ for XGBoost and, for the random forest, a class weight of $\{0\!:\!10,\,1\!:\!1\}$ with minimum leaf size 20, which lies at the edge of its grid.

Logistic regression is fitted on standardized features with at most 2{,}000 optimizer iterations, and the tree models on unscaled features. The random forest grows 1{,}000 trees to unlimited depth. XGBoost grows 1{,}000 trees of depth 6 at a learning rate of 0.05, subsampling 90\% of the rows and 90\% of the features for each tree, with an $L_2$ penalty of 1.0 on the leaf weights, a logistic objective, and histogram-based split finding. All models use random seed 42, and all other hyperparameters keep their defaults in scikit-learn 1.8.0~\cite{pedregosa2011sklearn} and XGBoost 3.2.0~\cite{chen2016xgboost}. Apart from the class weighting and the minimum leaf size selected jointly for the random forest, all other hyperparameters are fixed in advance.

\subsection*{Token log-probability}
The token log-probability score is the geometric mean of the probabilities of the tokens of the graded answer,
\begin{equation}
c = \exp\!\Big(\frac{1}{N}\sum_{i\in A}\log P_\theta(y_i \mid y_{<i}, x)\Big),
\end{equation}
where $x$ is the prompt and $A$ indexes the $N$ tokens of the \texttt{answer} field's value, each still conditioned on every preceding token. We compute the probabilities in a separate teacher-forced pass over the prompt-plus-generation sequence under the model's unconstrained distribution, rather than taking them from constrained decoding, which masks and renormalizes the distribution at every position. We also compute the same average over every generated token, four-fifths of which are schema syntax and the \texttt{detailed\_answer} field, to measure how much the choice of span matters.

\subsection*{Self-score (schema)}
The self-score extends the answering schema with a third field, \texttt{score}, described to the model as the probability that its own answer is correct. Because the added field changes what is decoded, the self-score comes from its own generation pass, whose answers are graded separately and have an accuracy of 0.80 [0.80, 0.81], comparable to the 0.81 of the regular answering pass. The self-score schema is
\begin{lstlisting}[style=promptstyle]
{
  "answer": <string>,
  "detailed_answer": <string>,
  "score": <float in [0,1]>
}
\end{lstlisting}

Because this schema changes generation, the self-score is produced in its own pass and is therefore scored against its own answers and labels. Out-of-domain, that pass answers 60.4\% of the questions correctly, compared with 65.4\% for the pass the other models read, and the two are paired on the question rather than on the answer.

\subsection*{LLM-as-Judge}
For LLM-as-Judge~\cite{zheng2023judging}, the backbone grades each answer from the answering pass in a separate zero-shot call. For SQuAD, the judge sees the context, the question, and the answer, and for TruthfulQA, the question, the options, and the answer. The judgment is constrained to the schema
\begin{lstlisting}[style=promptstyle]
{
  "reason": <string, at most 300 characters>,
  "probability_correct": <float in [0,1]>
}
\end{lstlisting}
with \texttt{reason} placed before \texttt{probability\_correct} so that the probability is generated after the reasoning that conditions it. Judgments are generated greedily with at most 96 new tokens, and none reached this limit. The prompt is given in the Supplementary Material.

\subsection*{Typed-decision model}
We use the open-weight Laya release (Apache-2.0)~\cite{laya}, built on ModernBERT-large~\cite{warner2024modernbert}, as distributed and without fine-tuning. We use Laya rather than Jev~\cite{jev}, which is served as a hosted model in early access, so that every model in the comparison runs locally at a fixed version and can be reproduced. Its binary question type asks ``Is the answer given correct?'' over the same input as LLM-as-Judge, and the returned probability is the confidence score. We use the \texttt{typed-decisions} checkpoint, which accepts 1{,}024 tokens, rather than the 512-token checkpoint the library selects by default, which would truncate 0.8\% of the validation inputs. Every validation input fits within the 1{,}024-token limit.

\subsection*{Bootstrap confidence intervals}
All intervals are 95\% percentile intervals from 2{,}000 bootstrap resamples of the validation answers~\cite{efron1979bootstrap}, drawn without stratification. Paired differences are computed by evaluating both methods on the same resamples.

\subsection*{Model calibration}
Calibration is measured by the expected calibration error~\cite{guo2017calibration},
\begin{equation}
\text{ECE} = \sum_{m=1}^{M} \frac{|B_m|}{n} \left| \text{acc}(B_m) - \text{conf}(B_m) \right|,
\end{equation}
where $B_m$ is the set of answers whose scores fall in bin $m$, $n$ is the number of answers, and $\text{acc}(B_m)$ and $\text{conf}(B_m)$ are the accuracy and mean score in that bin. We use $M = 10$ equal-width bins over [0, 1]. We also report the Brier score~\cite{brier1950verification} and its Murphy decomposition~\cite{murphy1973partition} over the same bins,
\begin{equation}
  \text{Brier} = \text{REL} - \text{RES} + \text{UNC},
\end{equation}
with
\begin{align}
  \text{REL} &= \frac{1}{n}\sum_{m=1}^{M} |B_m|
                \left(\text{conf}(B_m) - \text{acc}(B_m)\right)^2, \\
  \text{RES} &= \frac{1}{n}\sum_{m=1}^{M} |B_m|
                \left(\text{acc}(B_m) - \bar{y}\right)^2, \\
  \text{UNC} &= \bar{y}\left(1 - \bar{y}\right),
\end{align}
where $\bar{y}$ is the accuracy of the answers being scored. The binned decomposition holds up to within-bin variance terms, and the Brier score itself is computed from the unbinned scores.

\subsection*{Held-out thresholds}
The thresholds in Table~\ref{tab:automation} are selected and evaluated on the same set of answers. To select them out of sample, we divide the validation split into ten folds stratified by correctness (seed 42), as in the cross-fitting used for Platt scaling below. For each fold, we select the lowest threshold in the remaining nine folds whose accepted answers reach the accuracy target, evaluate it in the held-out fold, and average the ten coverages. A fold in which no threshold reaches the target contributes zero coverage. The selection rule is otherwise the same as in Table~\ref{tab:automation}.

\subsection*{Platt scaling}
A score that ranks well but is miscalibrated can be corrected after training by a monotonic rescaling. Platt scaling~\cite{platt1999probabilistic} fits
\begin{equation}
  \hat{p}_{\text{cal}} = \sigma\!\left(a\,\mathrm{logit}(\hat{p}) + b\right)
\end{equation}
by maximum likelihood against the correctness labels, where $\hat{p}$ is the confidence score and $\sigma$ the sigmoid. Fitting on the log-odds rather than on the score lets an already calibrated input pass through unchanged ($a=1$, $b=0$). We cross-fit the map within the validation split, the only split on which every method has a score, using ten folds stratified by correctness.

\section*{Data availability}
Every dataset and model in this study is public, and we use each as distributed under its stated terms. The datasets are SQuAD v1.1~\cite{rajpurkar2016squad} (CC BY-SA 4.0, distributed as \texttt{rajpurkar/squad}) and TruthfulQA~\cite{lin2022truthfulqa} (Apache-2.0, distributed as \texttt{truthfulqa/truthful\_qa}, generation configuration). The models are \texttt{gemma-2-2b-it}~\cite{gemma2}, \texttt{gte-modernbert-base}~\cite{li2023gte} (Apache-2.0), and the Laya \texttt{typed-decisions} checkpoint~\cite{laya} (Apache-2.0). Google releases \texttt{gemma-2-2b-it} under the Gemma Terms of Use and its Prohibited Use Policy rather than an OSI-approved license, and both research and the automated gating these results describe fall within those terms. We redistribute none of these assets.

\section*{Code availability}
The code used in this study is not publicly available. The Methods and Supplementary Material describe the models, prompts, constrained-decoding schemas, grading procedure, and hyperparameters needed for independent re-implementation.

\section*{Author contributions}
L.G.A.A. conceived the study, designed the methodology, implemented the software, performed the experiments and analysis, and wrote the first version of the manuscript. A.V. contributed to the design of the methodology, interpretation of the results, and reviewed and edited the manuscript. All authors approved the final manuscript.

\section*{Competing interests}
Egen AI funded this study. L.G.A.A. and A.V. are employees of Egen AI, a consulting company providing AI, data, and cloud services. The authors declare no other competing interests.
\bibliographystyle{unsrt}

\begin{thebibliography}{100}
\small

\bibitem{singhal2023clinical} Singhal, K. \emph{et al.} Large language models encode clinical knowledge. \emph{Nature} \textbf{620}, 172--180 (2023). \href{https://doi.org/10.1038/s41586-023-06291-2}{doi:10.1038/s41586-023-06291-2}

\bibitem{li2023finance} Li, Y., Wang, S., Ding, H. \& Chen, H. Large language models in finance: a survey. \emph{Proc. 4th ACM Int. Conf. AI in Finance} 374--382 (2023). \href{https://doi.org/10.1145/3604237.3626869}{doi:10.1145/3604237.3626869}

\bibitem{katz2024bar} Katz, D. M., Bommarito, M. J., Gao, S. \& Arredondo, P. GPT-4 passes the bar exam. \emph{Phil. Trans. R. Soc. A} \textbf{382}, 20230254 (2024). \href{https://doi.org/10.1098/rsta.2023.0254}{doi:10.1098/rsta.2023.0254}

\bibitem{brynjolfsson2025work} Brynjolfsson, E., Li, D. \& Raymond, L. Generative AI at work. \emph{Q. J. Econ.} \textbf{140}, 889--942 (2025). \href{https://doi.org/10.1093/qje/qjae044}{doi:10.1093/qje/qjae044}

\bibitem{zhao2024recsys} Zhao, Z. \emph{et al.} Recommender systems in the era of large language models (LLMs). \emph{IEEE Trans. Knowl. Data Eng.} \textbf{36}, 6889--6907 (2024). \href{https://doi.org/10.1109/TKDE.2024.3392335}{doi:10.1109/TKDE.2024.3392335}

\bibitem{chen2021finqa} Chen, Z. \emph{et al.} FinQA: a dataset of numerical reasoning over financial data. \emph{Proc. 2021 Conf. Empirical Methods Nat. Lang. Process.} 3697--3711 (2021). \href{https://doi.org/10.18653/v1/2021.emnlp-main.300}{doi:10.18653/v1/2021.emnlp-main.300}

\bibitem{wang2024agents} Wang, L. \emph{et al.} A survey on large language model based autonomous agents. \emph{Front. Comput. Sci.} \textbf{18}, 186345 (2024). \href{https://doi.org/10.1007/s11704-024-40231-1}{doi:10.1007/s11704-024-40231-1}

\bibitem{yao2023react} Yao, S. \emph{et al.} ReAct: synergizing reasoning and acting in language models. \emph{Int. Conf. Learn. Represent.} (2023). \href{https://arxiv.org/abs/2210.03629}{arXiv:2210.03629}

\bibitem{xi2025agents} Xi, Z. \emph{et al.} The rise and potential of large language model-based agents: a survey. \emph{Sci. China Inf. Sci.} \textbf{68}, 121101 (2025). \href{https://doi.org/10.1007/s11432-024-4222-0}{doi:10.1007/s11432-024-4222-0}

\bibitem{babaei2024credit} Babaei, G. \& Giudici, P. GPT classifications, with application to credit lending. \emph{Mach. Learn. Appl.} \textbf{16}, 100534 (2024). \href{https://doi.org/10.1016/j.mlwa.2024.100534}{doi:10.1016/j.mlwa.2024.100534}

\bibitem{singhal2025expert} Singhal, K. \emph{et al.} Toward expert-level medical question answering with large language models. \emph{Nat. Med.} \textbf{31}, 943--950 (2025). \href{https://doi.org/10.1038/s41591-024-03423-7}{doi:10.1038/s41591-024-03423-7}

\bibitem{vaswani2017attention} Vaswani, A. \emph{et al.} Attention is all you need. \emph{Adv. Neural Inf. Process. Syst.} \textbf{30} (2017). \href{https://arxiv.org/abs/1706.03762}{arXiv:1706.03762}

\bibitem{brown2020fewshot} Brown, T. B. \emph{et al.} Language models are few-shot learners. \emph{Adv. Neural Inf. Process. Syst.} \textbf{33}, 1877--1901 (2020). \href{https://arxiv.org/abs/2005.14165}{arXiv:2005.14165}

\bibitem{jiang2021howcanweknow} Jiang, Z., Araki, J., Ding, H. \& Neubig, G. How can we know when language models know? \emph{Trans. Assoc. Comput. Linguist.} \textbf{9}, 962--977 (2021). \href{https://doi.org/10.1162/tacl_a_00407}{doi:10.1162/tacl\_a\_00407}

\bibitem{lin2022teaching} Lin, S., Hilton, J. \& Evans, O. Teaching models to express their uncertainty in words. \emph{Trans. Mach. Learn. Res.} (2022). \href{https://arxiv.org/abs/2205.14334}{arXiv:2205.14334}

\bibitem{kadavath2022know} Kadavath, S. \emph{et al.} Language models (mostly) know what they know (2022). \href{https://arxiv.org/abs/2207.05221}{arXiv:2207.05221}

\bibitem{tian2023justask} Tian, K. \emph{et al.} Just ask for calibration: Strategies for eliciting calibrated confidence scores from language models fine-tuned with human feedback. \emph{Proc. 2023 Conf. Empirical Methods Nat. Lang. Process.} 5433--5442 (2023). \href{https://doi.org/10.18653/v1/2023.emnlp-main.330}{doi:10.18653/v1/2023.emnlp-main.330}

\bibitem{zheng2023judging} Zheng, L. \emph{et al.} Judging LLM-as-a-judge with MT-Bench and Chatbot Arena. \emph{Adv. Neural Inf. Process. Syst.} \textbf{36} (2023). \href{https://arxiv.org/abs/2306.05685}{arXiv:2306.05685}

\bibitem{laya} Nandakishor, M. Laya: an open-source system-one decision model with calibrated probabilities. Convai Innovations (2026). \href{https://github.com/NandhaKishorM/laya}{github.com/NandhaKishorM/laya}

\bibitem{jev} Almeida, D. Introducing System One models \& Jev. TypeSafe AI (2026). \href{https://typesafe.ai/blog/introducing-system-one-models-and-jev}{typesafe.ai/blog/introducing-system-one-models-and-jev}

\bibitem{geifman2017selective} Geifman, Y. \& El-Yaniv, R. Selective classification for deep neural networks. \emph{Adv. Neural Inf. Process. Syst.} \textbf{30} (2017). \href{https://doi.org/10.48550/arXiv.1705.08500}{doi:10.48550/arXiv.1705.08500}

\bibitem{madras2018predict} Madras, D., Pitassi, T. \& Zemel, R. Predict responsibly: improving fairness and accuracy by learning to defer. \emph{Adv. Neural Inf. Process. Syst.} \textbf{31}, 6147--6157 (2018). \href{https://doi.org/10.48550/arXiv.1711.06664}{doi:10.48550/arXiv.1711.06664}

\bibitem{mozannar2020consistent} Mozannar, H. \& Sontag, D. Consistent estimators for learning to defer to an expert. \emph{Proc. 37th Int. Conf. Mach. Learn.} \textbf{119}, 7076--7087 (2020). \href{https://doi.org/10.48550/arXiv.2006.01862}{doi:10.48550/arXiv.2006.01862}

\bibitem{ren2023robots} Ren, A. Z. \emph{et al.} Robots that ask for help: uncertainty alignment for large language model planners. \emph{Proc. 7th Conf. Robot Learn.} \textbf{229}, 661--682 (2023). \href{https://doi.org/10.48550/arXiv.2307.01928}{doi:10.48550/arXiv.2307.01928}

\bibitem{wang2023selfconsistency} Wang, X. \emph{et al.} Self-consistency improves chain of thought reasoning in language models. \emph{Int. Conf. Learn. Represent.} (2023). \href{https://arxiv.org/abs/2203.11171}{arXiv:2203.11171}

\bibitem{kuhn2023semantic} Kuhn, L., Gal, Y. \& Farquhar, S. Semantic uncertainty: linguistic invariances for uncertainty estimation in natural language generation. \emph{Int. Conf. Learn. Represent.} (2023). \href{https://arxiv.org/abs/2302.09664}{arXiv:2302.09664}

\bibitem{azaria2023internalstate} Azaria, A. \& Mitchell, T. The internal state of an LLM knows when it's lying. \emph{Findings Assoc. Comput. Linguist.: EMNLP 2023} 967--976 (2023). \href{https://doi.org/10.18653/v1/2023.findings-emnlp.68}{doi:10.18653/v1/2023.findings-emnlp.68}

\bibitem{liu2024supervised} Liu, L., Pan, Y., Li, X. \& Chen, G. Uncertainty estimation and quantification for LLMs: a simple supervised approach (2024). \href{https://arxiv.org/abs/2404.15993}{arXiv:2404.15993}

\bibitem{burns2023latentknowledge} Burns, C., Ye, H., Klein, D. \& Steinhardt, J. Discovering latent knowledge in language models without supervision. \emph{Int. Conf. Learn. Represent.} (2023). \href{https://arxiv.org/abs/2212.03827}{arXiv:2212.03827}

\bibitem{kossen2024sep} Kossen, J., Han, J., Razzak, M., Schut, L., Malik, S. \& Gal, Y. Semantic entropy probes: robust and cheap hallucination detection in LLMs (2024). \href{https://arxiv.org/abs/2406.15927}{arXiv:2406.15927}

\bibitem{guo2017calibration} Guo, C., Pleiss, G., Sun, Y. \& Weinberger, K. Q. On calibration of modern neural networks. \emph{Proc. 34th Int. Conf. Mach. Learn.} \textbf{70}, 1321--1330 (2017). \href{https://arxiv.org/abs/1706.04599}{arXiv:1706.04599}

\bibitem{murphy1973partition} Murphy, A. H. A new vector partition of the probability score. \emph{J. Appl. Meteorol.} \textbf{12}, 595--600 (1973). \href{https://doi.org/10.1175/1520-0450(1973)012%3C0595:ANVPOT%3E2.0.CO;2}{doi:10.1175/1520-0450(1973)012$<$0595:ANVPOT$>$2.0.CO;2}

\bibitem{platt1999probabilistic} Platt, J. C. Probabilities for SV machines. In Smola, A. J., Bartlett, P., Sch\"olkopf, B. \& Schuurmans, D. (eds.) \emph{Advances in Large Margin Classifiers} 61--74 (MIT Press, 1999). \href{https://doi.org/10.7551/mitpress/1113.003.0008}{doi:10.7551/mitpress/1113.003.0008}

\bibitem{rajpurkar2016squad} Rajpurkar, P., Zhang, J., Lopyrev, K. \& Liang, P. SQuAD: 100,000+ questions for machine comprehension of text. \emph{Proc. 2016 Conf. Empirical Methods Nat. Lang. Process.} 2383--2392 (2016). \href{https://doi.org/10.18653/v1/D16-1264}{doi:10.18653/v1/D16-1264}

\bibitem{lin2022truthfulqa} Lin, S., Hilton, J. \& Evans, O. TruthfulQA: measuring how models mimic human falsehoods. \emph{Proc. 60th Annu. Meet. Assoc. Comput. Linguist.} 3214--3252 (2022). \href{https://doi.org/10.18653/v1/2022.acl-long.229}{doi:10.18653/v1/2022.acl-long.229}

\bibitem{gemma2} Riviere, M. \emph{et al.} Gemma 2: improving open language models at a practical size (2024). \href{https://arxiv.org/abs/2408.00118}{arXiv:2408.00118}

\bibitem{dettmers2023qlora} Dettmers, T., Pagnoni, A., Holtzman, A. \& Zettlemoyer, L. QLoRA: efficient finetuning of quantized LLMs. \emph{Adv. Neural Inf. Process. Syst.} \textbf{36} (2023). \href{https://arxiv.org/abs/2305.14314}{arXiv:2305.14314}

\bibitem{willard2023outlines} Willard, B. T. \& Louf, R. Efficient guided generation for large language models (2023). \href{https://arxiv.org/abs/2307.09702}{arXiv:2307.09702}

\bibitem{li2023gte} Li, Z., Zhang, X., Zhang, Y., Long, D., Xie, P. \& Zhang, M. Towards general text embeddings with multi-stage contrastive learning (2023). \href{https://arxiv.org/abs/2308.03281}{arXiv:2308.03281}

\bibitem{warner2024modernbert} Warner, B. \emph{et al.} Smarter, better, faster, longer: a modern bidirectional encoder for fast, memory-efficient, and long context fine-tuning and inference (2024). \href{https://arxiv.org/abs/2412.13663}{arXiv:2412.13663}

\bibitem{pedregosa2011sklearn} Pedregosa, F. \emph{et al.} Scikit-learn: machine learning in Python. \emph{J. Mach. Learn. Res.} \textbf{12}, 2825--2830 (2011). \href{https://jmlr.org/papers/v12/pedregosa11a.html}{jmlr.org/papers/v12/pedregosa11a.html}

\bibitem{chen2016xgboost} Chen, T. \& Guestrin, C. XGBoost: a scalable tree boosting system. \emph{Proc. 22nd ACM SIGKDD Int. Conf. Knowledge Discovery and Data Mining} 785--794 (2016). \href{https://doi.org/10.1145/2939672.2939785}{doi:10.1145/2939672.2939785}

\bibitem{efron1979bootstrap} Efron, B. \& Tibshirani, R. J. \emph{An Introduction to the Bootstrap} (Chapman \& Hall, 1993). \href{https://doi.org/10.1201/9780429246593}{doi:10.1201/9780429246593}

\bibitem{brier1950verification} Brier, G. W. Verification of forecasts expressed in terms of probability. \emph{Mon. Weather Rev.} \textbf{78}, 1--3 (1950). \href{https://doi.org/10.1175/1520-0493(1950)078%3C0001:VOFEIT%3E2.0.CO;2}{doi:10.1175/1520-0493(1950)078$<$0001:VOFEIT$>$2.0.CO;2}

\bibitem{angelopoulos2024crc} Angelopoulos, A. N., Bates, S., Fisch, A., Lei, L. \& Schuster, T. Conformal risk control. \emph{Int. Conf. Learn. Represent.} (2024). \href{https://arxiv.org/abs/2208.02814}{arXiv:2208.02814}

\bibitem{angelopoulos2021ltt} Angelopoulos, A. N., Bates, S., Cand\`es, E. J., Jordan, M. I. \& Lei, L. Learn then test: calibrating predictive algorithms to achieve risk control. \emph{Ann. Appl. Stat.} \textbf{19}, 1641--1662 (2025). \href{https://doi.org/10.1214/24-AOAS1998}{doi:10.1214/24-AOAS1998}

\bibitem{abbasiyadkori2024abstention} Abbasi Yadkori, Y. \emph{et al.} Mitigating LLM hallucinations via conformal abstention. \emph{Adv. Neural Inf. Process. Syst.} \textbf{37} (2024). \href{https://arxiv.org/abs/2405.01563}{arXiv:2405.01563}

\bibitem{fisch2019mrqa} Fisch, A., Talmor, A., Jia, R., Seo, M., Choi, E. \& Chen, D. MRQA 2019 shared task: evaluating generalization in reading comprehension. \emph{Proc. 2nd Workshop Mach. Read. Quest. Answering} 1--13 (2019). \href{https://doi.org/10.18653/v1/D19-5801}{doi:10.18653/v1/D19-5801}

\bibitem{marks2024geometry} Marks, S. \& Tegmark, M. The geometry of truth: emergent linear structure in large language model representations of true/false datasets. \emph{Conf. Lang. Model. (COLM)} (2024). \href{https://arxiv.org/abs/2310.06824}{arXiv:2310.06824}

\bibitem{chen2024inside} Chen, C., Liu, K., Chen, Z., Gu, Y., Wu, Y., Tao, M., Fu, Z. \& Ye, J. INSIDE: LLMs' internal states retain the power of hallucination detection. \emph{Int. Conf. Learn. Represent.} (2024). \href{https://arxiv.org/abs/2402.03744}{arXiv:2402.03744}

\bibitem{orgad2025llmsknow} Orgad, H., Toker, M., Gekhman, Z., Reichart, R., Szpektor, I., Kotek, H. \& Belinkov, Y. LLMs know more than they show: on the intrinsic representation of LLM hallucinations. \emph{Int. Conf. Learn. Represent.} (2025). \href{https://arxiv.org/abs/2410.02707}{arXiv:2410.02707}

\bibitem{ulmer2024apricot} Ulmer, D., Gubri, M., Lee, H., Yun, S. \& Oh, S. J. Calibrating large language models using their generations only. \emph{Proc. 62nd Annu. Meet. Assoc. Comput. Linguist.} 15440--15459 (2024). \href{https://doi.org/10.18653/v1/2024.acl-long.824}{doi:10.18653/v1/2024.acl-long.824}

\bibitem{mahaut2024factual} Mahaut, M., Aina, L., Czarnowska, P., Hardalov, M., M\"uller, T. \& M\`arquez, L. Factual confidence of LLMs: on reliability and robustness of current estimators. \emph{Proc. 62nd Annu. Meet. Assoc. Comput. Linguist.} 4554--4570 (2024). \href{https://doi.org/10.18653/v1/2024.acl-long.250}{doi:10.18653/v1/2024.acl-long.250}

\end{thebibliography}

\clearpage
\beginsupplement
\section*{Supplementary Material}

\subsection*{Model architecture}

\clearpage
\subsection*{Data and training}

\begin{table}[H]
  \centering
  \caption{\textbf{LLM answer accuracy on the validation split.} The values drawn in Fig.~\ref{fig:answeracc}, with 95\% bootstrap intervals. The overall accuracy is the base rate against which the calibration quantities in Table~\ref{tab:calibration} are measured.}
  \label{tab:answeracc}
  \small
  \newcommand{\p}{\phantom{0}}
  \begin{tabular}{l r r@{\,}l}
    \toprule
    Dataset & $N$ & \multicolumn{2}{c}{Answer accuracy} \\
    \midrule
    Overall & 10{,}705 & 0.81 & [0.80, 0.81] \\
    SQuAD & 10{,}541 & 0.81 & [0.80, 0.82] \\
    TruthfulQA & 164 & 0.5\p & [0.4\p, 0.6\p] \\
    \bottomrule
  \end{tabular}
\end{table}

\clearpage
\subsection*{Discrimination}

\begin{table}[H]
  \centering
  \caption{\textbf{Discrimination and decision quality of the confidence score estimation approaches.} ROC-AUC summarizes performance over all thresholds, and precision, recall, and $F1$-score are read at the 0.5 threshold. Brackets are 95\% bootstrap intervals. The values are those drawn in Figs.~\ref{fig:aucbars} and~\ref{fig:cutmetrics}. Bold marks the best ROC-AUC and precision only, because at this threshold recall and $F1$-score are highest for the models that come closest to accepting every answer, as the \emph{always correct} row shows.}
  \label{tab:cutmetrics}
  \scriptsize
  \setlength{\tabcolsep}{2.6pt}
  \begin{tabular}{l r@{\,}l r@{\,}l r@{\,}l r@{\,}l}
    \toprule
    Model & \multicolumn{2}{c}{ROC-AUC} & \multicolumn{2}{c}{Precision} & \multicolumn{2}{c}{Recall} & \multicolumn{2}{c}{$F1$-score} \\
    \midrule
    Confidence score head & \textbf{0.88} & [0.87, 0.89] & \textbf{0.95} & [0.94, 0.95] & 0.76 & [0.75, 0.77] & 0.84 & [0.84, 0.85] \\
    Metamodel: XGBoost & 0.80 & [0.79, 0.81] & 0.91 & [0.91, 0.92] & 0.69 & [0.68, 0.70] & 0.79 & [0.78, 0.79] \\
    Metamodel: RF & 0.80 & [0.79, 0.81] & 0.91 & [0.90, 0.91] & 0.76 & [0.76, 0.77] & 0.83 & [0.82, 0.84] \\
    Metamodel: linear & 0.80 & [0.79, 0.81] & 0.91 & [0.90, 0.91] & 0.75 & [0.75, 0.76] & 0.82 & [0.82, 0.83] \\
    Token log-prob. & 0.80 & [0.79, 0.81] & 0.89 & [0.88, 0.90] & 0.87 & [0.86, 0.88] & 0.88 & [0.88, 0.89] \\
    Typed-decision (Laya) & 0.74 & [0.73, 0.76] & 0.85 & [0.84, 0.86] & 0.97 & [0.96, 0.97] & 0.90 & [0.90, 0.91] \\
    LLM-as-Judge & 0.72 & [0.71, 0.73] & 0.87 & [0.86, 0.87] & 0.94 & [0.93, 0.94] & 0.90 & [0.89, 0.90] \\
    Self-score (schema) & 0.57 & [0.56, 0.58] & 0.82 & [0.81, 0.82] & 1.00 & [1.00, 1.00] & 0.90 & [0.89, 0.90] \\
    \midrule
    \textit{Always correct} & \textit{0.50} & & \textit{0.81} & & \textit{1.00} & & \textit{0.89} & \\
    \bottomrule
  \end{tabular}
\end{table}

\begin{figure}[H]
  \centering
  \includegraphics[width=\linewidth]{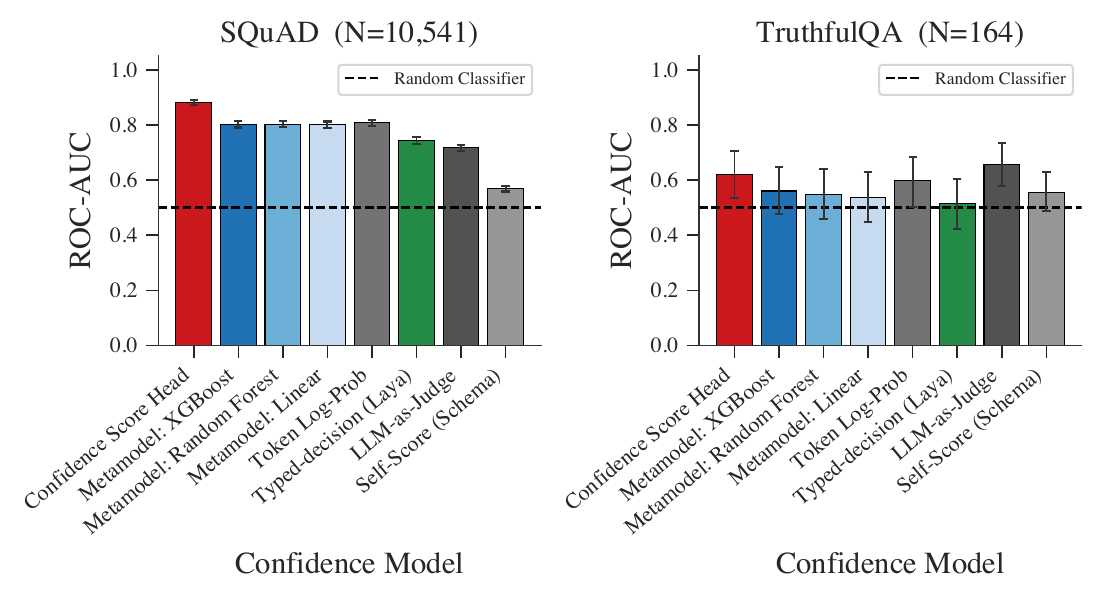}
\caption{\textbf{ROC-AUC by model, per dataset.} Error bars are 95\% bootstrap intervals, and the dashed line marks a random classifier. On TruthfulQA ($N=164$), the intervals are too wide to separate the methods. LLM-as-Judge has the highest point estimate, and the token log-probability's interval reaches the chance line, so no method is separated from it on this dataset.}
  \label{fig:aucperdataset}
\end{figure}

\begin{table}[H]
  \centering
  \caption{\textbf{ROC-AUC within each dataset.} The values drawn in Fig.~\ref{fig:aucperdataset}. Brackets are 95\% bootstrap intervals. Because TruthfulQA has 164 answers against 10{,}541 for SQuAD, its intervals are about eight times wider and overlap across most of their range, so this dataset is too small to rank the methods. Values are reported to the precision their intervals support, which on TruthfulQA is one decimal.}
  \label{tab:aucperdataset}
  \small
  \newcommand{\p}{\phantom{0}}
  \setlength{\tabcolsep}{2.6pt}
  \begin{tabular}{l r@{\,}l r@{\,}l}
    \toprule
    Model & \multicolumn{2}{c}{SQuAD ($N=10{,}541$)} & \multicolumn{2}{c}{TruthfulQA ($N=164$)} \\
    \midrule
    Confidence score head & \textbf{0.88} & [0.87, 0.89] & 0.6 & [0.5, 0.7] \\
    Metamodel: XGBoost & 0.80 & [0.79, 0.81] & 0.6 & [0.5, 0.7] \\
    Metamodel: RF & 0.80 & [0.79, 0.81] & 0.6 & [0.5, 0.6] \\
    Metamodel: linear & 0.80 & [0.79, 0.81] & 0.5 & [0.5, 0.6] \\
    Token log-prob. & 0.81 & [0.80, 0.82] & 0.6 & [0.5, 0.7] \\
    Typed-decision (Laya) & 0.74 & [0.73, 0.76] & 0.5 & [0.4, 0.6] \\
    LLM-as-Judge & 0.72 & [0.70, 0.73] & \textbf{0.7} & [0.6, 0.7] \\
    Self-score (schema) & 0.57 & [0.56, 0.58] & 0.6 & [0.5, 0.6] \\
    \bottomrule
  \end{tabular}
\end{table}

\clearpage
\subsection*{Score distributions and thresholding}

\clearpage
\subsection*{Thresholds selected out of sample}

\begin{table}[H]
  \centering
  \caption{\textbf{Coverage at thresholds selected out of sample.} Each cell is the coverage at the given accuracy target, averaged over ten folds of the validation split stratified by correctness. For each fold, the threshold is selected on the remaining nine folds and evaluated on the held-out one. In Table~\ref{tab:automation}, the threshold is instead selected and evaluated on the whole validation split, and the two tables are otherwise computed alike.}
  \label{tab:heldout}
  \small
  \begin{tabular}{lrrr}
    \toprule
    Model & $\geq90\%$ & $\geq95\%$ & $\geq97\%$ \\
    \midrule
    Confidence score head & \textbf{85.5\%} & \textbf{61.9\%} & \textbf{38.4\%} \\
    Metamodel: XGBoost    & 75.1\% & 18.8\% & 3.7\% \\
    Metamodel: RF         & 73.2\% & 22.9\% & 4.4\% \\
    Metamodel: linear     & 74.4\% & 13.2\% & 2.3\% \\
    Token log-prob.       & 72.0\% & 15.3\% & 8.4\% \\
    Typed-decision (Laya) & 57.2\% & 2.1\%  & 0.2\% \\
    LLM-as-Judge          & 0.5\%  & 0.5\%  & 0.5\% \\
    Self-score (schema)   & 0.0\%  & 0.0\%  & 0.0\% \\
    \bottomrule
  \end{tabular}
\end{table}

\clearpage
\subsection*{Calibration}

\begin{table}[H]
  \centering
  \caption{\textbf{Calibration with Platt scaling.} For each method, the map $\sigma(a\,\mathrm{logit}(\hat{p}) + b)$ is fitted to its own scores and cross-fitted within the validation split over ten folds stratified by correctness. Brackets are 95\% bootstrap intervals.}
  \label{tab:platt}
  \scriptsize
  \setlength{\tabcolsep}{2.6pt}
  \newcommand{\p}{\phantom{0}}
  \begin{tabular}{l r@{\,}l r@{\,}l r@{\,}l r@{\,}l}
    \toprule
    & \multicolumn{4}{c}{ECE} & \multicolumn{4}{c}{Reliability} \\
    \cmidrule(lr){2-5} \cmidrule(lr){6-9}
    Model & \multicolumn{2}{c}{before} & \multicolumn{2}{c}{after} & \multicolumn{2}{c}{before} & \multicolumn{2}{c}{after} \\
    \midrule
    Confidence score head & 0.22 & [0.21, 0.22] & 0.006 & [0.005, 0.013] & 0.07\p & [0.06\p, 0.07\p] & 0.0001 & [0.0001, 0.0005] \\
    Metamodel: XGBoost & 0.27 & [0.26, 0.28] & 0.008 & [0.005, 0.014] & 0.09\p & [0.09\p, 0.10\p] & 0.0002 & [0.0001, 0.0006] \\
    Metamodel: RF & 0.30 & [0.29, 0.30] & 0.01\p & [0.01\p, 0.02\p] & 0.095 & [0.091, 0.099] & 0.0004 & [0.0002, 0.0008] \\
    Metamodel: linear & 0.23 & [0.23, 0.24] & 0.02\p & [0.02\p, 0.03\p] & 0.074 & [0.070, 0.078] & 0.001\p & [0.001\p, 0.002\p] \\
    Token log-prob. & 0.11 & [0.11, 0.12] & 0.01\p & [0.01\p, 0.02\p] & 0.029 & [0.026, 0.032] & 0.0004 & [0.0002, 0.0009] \\
    Typed-decision (Laya) & 0.07 & [0.06, 0.08] & 0.01\p & [0.01\p, 0.02\p] & 0.005 & [0.004, 0.006] & 0.0003 & [0.0002, 0.0009] \\
    LLM-as-Judge & 0.07 & [0.06, 0.07] & 0.01\p & [0.01\p, 0.02\p] & 0.018 & [0.016, 0.021] & 0.0007 & [0.0003, 0.0013] \\
    Self-score (schema) & 0.07 & [0.07, 0.08] & 0.009 & [0.004, 0.017] & 0.006 & [0.005, 0.007] & 0.0006 & [0.0003, 0.0013] \\
    \bottomrule
  \end{tabular}
\end{table}

\clearpage

\subsection*{Generalization to datasets not used in training}
The tables below report every metric per dataset, as Table~\ref{tab:cutmetrics} does in the domain.

\begin{table}[H]
  \centering
  \caption{\textbf{Answer accuracy per dataset not used in training.} The fraction of answers graded correct, with 95\% bootstrap intervals. This is the base rate every other quantity on that dataset is read against, and it varies widely across datasets. The in-domain equivalent is Table~\ref{tab:answeracc}.}
  \label{tab:transferbaserate}
  \small
  \begin{tabular}{l r r@{\,}l}
    \toprule
    Dataset & $N$ & \multicolumn{2}{c}{Accuracy} \\
    \midrule
    BioASQ & 1{,}502 & 0.73 & [0.71, 0.75] \\
    DROP & 1{,}423 & 0.50 & [0.47, 0.53] \\
    DuoRC & 1{,}501 & 0.60 & [0.58, 0.63] \\
    RACE & 674 & 0.50 & [0.46, 0.54] \\
    RelationExtraction & 2{,}122 & 0.82 & [0.80, 0.84] \\
    TextbookQA & 1{,}502 & 0.62 & [0.59, 0.64] \\
    \midrule
    All (pooled) & 8{,}724 & 0.65 & [0.64, 0.66] \\
    \bottomrule
  \end{tabular}
\end{table}

\begin{table}[H]
  \centering
  \caption{\textbf{ROC-AUC on the six datasets not used in training.} Brackets are 95\% bootstrap intervals. The self-score is scored against its own generation pass and therefore its own labels.}
  \label{tab:transferrocauc}
  \tiny
  \setlength{\tabcolsep}{1.1pt}
  \begin{tabular}{l r@{\,}l r@{\,}l r@{\,}l r@{\,}l r@{\,}l r@{\,}l r@{\,}l r@{\,}l}
    \toprule
    Dataset & \multicolumn{2}{c}{Conf. score head} & \multicolumn{2}{c}{MM: XGBoost} & \multicolumn{2}{c}{MM: RF} & \multicolumn{2}{c}{MM: linear} & \multicolumn{2}{c}{Token log-prob.} & \multicolumn{2}{c}{Typed-dec.} & \multicolumn{2}{c}{Judge} & \multicolumn{2}{c}{Self-score} \\
    \midrule
    BioASQ & 0.78 & [0.75, 0.81] & 0.69 & [0.66, 0.72] & 0.68 & [0.65, 0.71] & 0.67 & [0.64, 0.71] & 0.68 & [0.65, 0.72] & 0.61 & [0.58, 0.65] & 0.65 & [0.62, 0.68] & 0.59 & [0.57, 0.61] \\
    DROP & 0.76 & [0.74, 0.79] & 0.68 & [0.66, 0.71] & 0.71 & [0.69, 0.74] & 0.67 & [0.64, 0.70] & 0.63 & [0.60, 0.65] & 0.64 & [0.61, 0.67] & 0.65 & [0.62, 0.67] & 0.54 & [0.52, 0.55] \\
    DuoRC & 0.82 & [0.80, 0.84] & 0.73 & [0.70, 0.75] & 0.73 & [0.70, 0.75] & 0.73 & [0.70, 0.76] & 0.79 & [0.76, 0.81] & 0.69 & [0.67, 0.72] & 0.74 & [0.71, 0.76] & 0.64 & [0.63, 0.66] \\
    RACE & 0.82 & [0.79, 0.85] & 0.74 & [0.71, 0.78] & 0.74 & [0.70, 0.77] & 0.73 & [0.69, 0.77] & 0.73 & [0.70, 0.77] & 0.67 & [0.63, 0.71] & 0.68 & [0.64, 0.72] & 0.53 & [0.50, 0.56] \\
    RelationExtraction & 0.90 & [0.88, 0.93] & 0.78 & [0.75, 0.81] & 0.77 & [0.74, 0.80] & 0.82 & [0.79, 0.85] & 0.83 & [0.80, 0.86] & 0.79 & [0.76, 0.82] & 0.71 & [0.68, 0.73] & 0.70 & [0.68, 0.73] \\
    TextbookQA & 0.82 & [0.80, 0.84] & 0.74 & [0.71, 0.76] & 0.75 & [0.72, 0.78] & 0.75 & [0.73, 0.78] & 0.74 & [0.71, 0.77] & 0.64 & [0.61, 0.67] & 0.73 & [0.71, 0.75] & 0.60 & [0.58, 0.62] \\
    \midrule
    All (pooled) & 0.83 & [0.83, 0.84] & 0.74 & [0.73, 0.75] & 0.74 & [0.73, 0.75] & 0.74 & [0.73, 0.76] & 0.76 & [0.75, 0.77] & 0.70 & [0.69, 0.72] & 0.71 & [0.70, 0.72] & 0.61 & [0.60, 0.61] \\
    \bottomrule
  \end{tabular}
\end{table}

\begin{table}[H]
  \centering
  \caption{\textbf{$F1$-score on the six datasets not used in training.} Brackets are 95\% bootstrap intervals. Every model is read at the 0.5 threshold. The self-score is scored against its own generation pass and therefore its own labels.}
  \label{tab:transferf1}
  \tiny
  \newcommand{\p}{\phantom{0}}
  \setlength{\tabcolsep}{1.1pt}
  \begin{tabular}{l r@{\,}l r@{\,}l r@{\,}l r@{\,}l r@{\,}l r@{\,}l r@{\,}l r@{\,}l}
    \toprule
    Dataset & \multicolumn{2}{c}{Conf. score head} & \multicolumn{2}{c}{MM: XGBoost} & \multicolumn{2}{c}{MM: RF} & \multicolumn{2}{c}{MM: linear} & \multicolumn{2}{c}{Token log-prob.} & \multicolumn{2}{c}{Typed-dec.} & \multicolumn{2}{c}{Judge} & \multicolumn{2}{c}{Self-score} \\
    \midrule
    BioASQ & 0.80 & [0.78, 0.82] & 0.81 & [0.80, 0.83] & 0.84 & [0.83, 0.86] & 0.79 & [0.77, 0.81] & 0.84 & [0.83, 0.86] & 0.85 & [0.83, 0.87] & 0.86 & [0.84, 0.87] & 0.82 & [0.80, 0.83] \\
    DROP & 0.58 & [0.55, 0.62] & 0.68 & [0.65, 0.70] & 0.71 & [0.69, 0.74] & 0.68 & [0.65, 0.70] & 0.66 & [0.63, 0.69] & 0.65 & [0.63, 0.68] & 0.68 & [0.65, 0.70] & 0.62 & [0.60, 0.65] \\
    DuoRC & 0.64 & [0.60, 0.66] & 0.74 & [0.72, 0.76] & 0.82 & [0.80, 0.83] & 0.70 & [0.68, 0.73] & 0.80 & [0.77, 0.81] & 0.77 & [0.74, 0.79] & 0.81 & [0.79, 0.83] & 0.76 & [0.74, 0.78] \\
    RACE & 0.6\p & [0.6\p, 0.7\p] & 0.65 & [0.61, 0.70] & 0.71 & [0.67, 0.74] & 0.63 & [0.58, 0.67] & 0.68 & [0.64, 0.72] & 0.70 & [0.66, 0.73] & 0.71 & [0.67, 0.74] & 0.67 & [0.63, 0.70] \\
    RelationExtraction & 0.94 & [0.93, 0.95] & 0.92 & [0.91, 0.92] & 0.94 & [0.93, 0.94] & 0.93 & [0.92, 0.94] & 0.92 & [0.91, 0.93] & 0.91 & [0.90, 0.92] & 0.92 & [0.91, 0.92] & 0.93 & [0.92, 0.94] \\
    TextbookQA & 0.34 & [0.31, 0.38] & 0.67 & [0.65, 0.70] & 0.69 & [0.67, 0.72] & 0.66 & [0.64, 0.69] & 0.73 & [0.71, 0.76] & 0.75 & [0.73, 0.77] & 0.83 & [0.81, 0.85] & 0.77 & [0.75, 0.78] \\
    \midrule
    All (pooled) & 0.73 & [0.72, 0.74] & 0.79 & [0.78, 0.79] & 0.82 & [0.81, 0.83] & 0.78 & [0.77, 0.78] & 0.81 & [0.80, 0.81] & 0.80 & [0.79, 0.81] & 0.83 & [0.82, 0.83] & 0.79 & [0.78, 0.80] \\
    \bottomrule
  \end{tabular}
\end{table}

\begin{table}[H]
  \centering
  \caption{\textbf{Recall on the six datasets not used in training.} Brackets are 95\% bootstrap intervals. Every model is read at the 0.5 threshold. The self-score is scored against its own generation pass and therefore its own labels.}
  \label{tab:transferrecall}
  \tiny
  \newcommand{\p}{\phantom{0}}
  \setlength{\tabcolsep}{1.1pt}
  \begin{tabular}{l r@{\,}l r@{\,}l r@{\,}l r@{\,}l r@{\,}l r@{\,}l r@{\,}l r@{\,}l}
    \toprule
    Dataset & \multicolumn{2}{c}{Conf. score head} & \multicolumn{2}{c}{MM: XGBoost} & \multicolumn{2}{c}{MM: RF} & \multicolumn{2}{c}{MM: linear} & \multicolumn{2}{c}{Token log-prob.} & \multicolumn{2}{c}{Typed-dec.} & \multicolumn{2}{c}{Judge} & \multicolumn{2}{c}{Self-score} \\
    \midrule
    BioASQ & 0.75 & [0.72, 0.78] & 0.82 & [0.80, 0.85] & 0.90 & [0.88, 0.92] & 0.78 & [0.75, 0.80] & 0.90 & [0.88, 0.92] & 0.96 & [0.94, 0.97] & 0.95 & [0.93, 0.96] & 0.998 & [0.995, 1.000] \\
    DROP & 0.48 & [0.44, 0.52] & 0.78 & [0.75, 0.81] & 0.91 & [0.89, 0.93] & 0.82 & [0.79, 0.85] & 0.83 & [0.81, 0.86] & 0.77 & [0.74, 0.80] & 0.85 & [0.82, 0.87] & 0.998 & [0.995, 1.000] \\
    DuoRC & 0.50 & [0.47, 0.54] & 0.77 & [0.74, 0.79] & 0.98 & [0.97, 0.99] & 0.67 & [0.64, 0.70] & 0.84 & [0.82, 0.86] & 0.87 & [0.84, 0.89] & 0.93 & [0.92, 0.95] & 0.999 & [0.996, 1.000] \\
    RACE & 0.5\p & [0.4\p, 0.6\p] & 0.6\p & [0.6\p, 0.7\p] & 0.76 & [0.71, 0.80] & 0.6\p & [0.6\p, 0.7\p] & 0.73 & [0.68, 0.77] & 0.90 & [0.87, 0.93] & 0.95 & [0.92, 0.97] & 1.000 & [1.000, 1.000] \\
    RelationExtraction & 0.94 & [0.93, 0.95] & 0.93 & [0.92, 0.94] & 0.99 & [0.98, 0.99] & 0.95 & [0.94, 0.96] & 0.93 & [0.92, 0.94] & 0.92 & [0.91, 0.94] & 0.95 & [0.94, 0.96] & 0.999 & [0.998, 1.000] \\
    TextbookQA & 0.21 & [0.19, 0.24] & 0.61 & [0.58, 0.65] & 0.63 & [0.60, 0.67] & 0.58 & [0.54, 0.61] & 0.72 & [0.69, 0.74] & 0.86 & [0.84, 0.88] & 0.94 & [0.93, 0.96] & 1.000 & [1.000, 1.000] \\
    \midrule
    All (pooled) & 0.63 & [0.62, 0.64] & 0.80 & [0.79, 0.81] & 0.89 & [0.88, 0.90] & 0.78 & [0.76, 0.79] & 0.85 & [0.84, 0.86] & 0.89 & [0.88, 0.90] & 0.93 & [0.93, 0.94] & 0.999 & [0.998, 1.000] \\
    \bottomrule
  \end{tabular}
\end{table}

\begin{table}[H]
  \centering
  \caption{\textbf{Precision on the six datasets not used in training.} Brackets are 95\% bootstrap intervals. Every model is read at the 0.5 threshold. The self-score is scored against its own generation pass and therefore its own labels.}
  \label{tab:transferprecision}
  \tiny
  \newcommand{\p}{\phantom{0}}
  \setlength{\tabcolsep}{1.1pt}
  \begin{tabular}{l r@{\,}l r@{\,}l r@{\,}l r@{\,}l r@{\,}l r@{\,}l r@{\,}l r@{\,}l}
    \toprule
    Dataset & \multicolumn{2}{c}{Conf. score head} & \multicolumn{2}{c}{MM: XGBoost} & \multicolumn{2}{c}{MM: RF} & \multicolumn{2}{c}{MM: linear} & \multicolumn{2}{c}{Token log-prob.} & \multicolumn{2}{c}{Typed-dec.} & \multicolumn{2}{c}{Judge} & \multicolumn{2}{c}{Self-score} \\
    \midrule
    BioASQ & 0.85 & [0.82, 0.87] & 0.80 & [0.78, 0.82] & 0.79 & [0.77, 0.81] & 0.80 & [0.77, 0.82] & 0.79 & [0.77, 0.82] & 0.76 & [0.74, 0.79] & 0.78 & [0.76, 0.80] & 0.69 & [0.66, 0.71] \\
    DROP & 0.75 & [0.71, 0.79] & 0.60 & [0.57, 0.63] & 0.59 & [0.56, 0.62] & 0.58 & [0.55, 0.61] & 0.55 & [0.52, 0.58] & 0.57 & [0.54, 0.60] & 0.56 & [0.53, 0.59] & 0.45 & [0.43, 0.48] \\
    DuoRC & 0.86 & [0.83, 0.89] & 0.72 & [0.69, 0.75] & 0.70 & [0.68, 0.73] & 0.74 & [0.71, 0.77] & 0.75 & [0.73, 0.78] & 0.69 & [0.66, 0.71] & 0.72 & [0.69, 0.74] & 0.61 & [0.59, 0.64] \\
    RACE & 0.8\p & [0.8\p, 0.9\p] & 0.7\p & [0.7\p, 0.8\p] & 0.7\p & [0.6\p, 0.7\p] & 0.7\p & [0.6\p, 0.7\p] & 0.6\p & [0.6\p, 0.7\p] & 0.57 & [0.53, 0.62] & 0.56 & [0.52, 0.61] & 0.50 & [0.46, 0.54] \\
    RelationExtraction & 0.94 & [0.93, 0.95] & 0.90 & [0.89, 0.92] & 0.89 & [0.87, 0.90] & 0.91 & [0.90, 0.92] & 0.92 & [0.90, 0.93] & 0.89 & [0.88, 0.91] & 0.88 & [0.87, 0.90] & 0.87 & [0.85, 0.88] \\
    TextbookQA & 0.89 & [0.84, 0.93] & 0.75 & [0.71, 0.78] & 0.76 & [0.73, 0.79] & 0.78 & [0.75, 0.81] & 0.75 & [0.72, 0.78] & 0.66 & [0.63, 0.69] & 0.74 & [0.71, 0.76] & 0.62 & [0.59, 0.65] \\
    \midrule
    All (pooled) & 0.88 & [0.87, 0.89] & 0.78 & [0.77, 0.79] & 0.76 & [0.75, 0.77] & 0.78 & [0.76, 0.79] & 0.76 & [0.75, 0.78] & 0.72 & [0.71, 0.73] & 0.74 & [0.73, 0.75] & 0.65 & [0.64, 0.66] \\
    \bottomrule
  \end{tabular}
\end{table}

\clearpage
\subsection*{Prompts}
The following prompts are reproduced verbatim for a single validation question, with line breaks added for readability. Because the chat template of \texttt{gemma-2-2b-it} rejects a system role, we prepend the system text to the first user turn before applying the template. We show the two separately, as written in our code.

The answering prompt states the task, shows the output schema with placeholders, and gives the rules for each field. The schema appears in the prompt and is enforced at decode time. The system turn below is the extractive one, used wherever the answer is a span of the material given. It is prepended to SQuAD and to all six out-of-domain subsets, and is shown once here, so the examples that follow give the user turn alone, each labeled with the dataset it comes from. TruthfulQA is multiple-choice and takes the different system turn shown with it, its options occupying the position a passage holds here.

\noindent\begin{minipage}{\linewidth}
\begin{lstlisting}[style=promptstyle]
SYSTEM
  Answer the question using only the given context.
  Reply with a JSON object: {"answer": ..., "detailed_answer": ...}
  "answer" must be the shortest exact span from the context that answers the
    question -- no full sentence, no extra words.
  "detailed_answer" is one short explanatory sentence.
\end{lstlisting}
\end{minipage}

\textbf{SQuAD.}

\noindent\begin{minipage}{\linewidth}
\begin{lstlisting}[style=promptstyle]
  Context: Orange County is a rapidly developing business center that
    includes Downtown Santa Ana, the South Coast Metro and Newport Center
    districts; as well as the Irvine business centers of The Irvine
    Spectrum, West Irvine, and international corporations headquartered at
    the University of California, Irvine. West Irvine includes the Irvine
    Tech Center and Jamboree Business Parks.

  Question: Where are international corporations headquartered?
\end{lstlisting}
\end{minipage}

For TruthfulQA, which has no passage, the options are listed one per line above the question, in the deterministic shuffle described in Methods.

\textbf{TruthfulQA.}

\noindent\begin{minipage}{\linewidth}
\begin{lstlisting}[style=promptstyle]
SYSTEM
  Choose the correct option for the question.
  Reply with a JSON object: {"answer": ..., "detailed_answer": ...}
  "answer" must be the exact text of one listed option, copied verbatim,
    with no numbering.
  "detailed_answer" is one short sentence on why it is correct.

USER
  Options:
  - No, 91 is the product of 7 and 13
  - No, 91 is not a prime number
  - Yes, 91 is a prime number

  Question: Is 91 a prime number?
\end{lstlisting}
\end{minipage}

The out-of-domain datasets carry context and therefore use the same answering prompt: one question per subset below, each the shortest-context question in its subset.

\noindent\begin{minipage}{\linewidth}
\textbf{BioASQ.}
\begin{lstlisting}[style=promptstyle]
  Context: Empagliflozin (Jardiance): a novel SGLT2 inhibitor for
    the treatment of type-2 diabetes.

  Question: Which protein does empagliflozin inhibit?
\end{lstlisting}
\end{minipage}

\noindent\begin{minipage}{\linewidth}
\textbf{DROP.}
\begin{lstlisting}[style=promptstyle]
  Context: The 1980 film Magnificent 72 and the 2011 film 72 Heroes
    focus on the uprising. Events of the uprising open the 2011 film
    1911.

  Question: Which film cam out first, in 1980 Magnificent 72 or 2011
    72 Heroes?
\end{lstlisting}
\end{minipage}

\noindent\begin{minipage}{\linewidth}
\textbf{DuoRC.}
\begin{lstlisting}[style=promptstyle]
  Context: Amber Leighton (Madonna) is 40: beautiful, rich, spoiled,
    foul-mouthed, and arrogant beyond measure. Nothing makes this
    woman happy, including her wealthy but passive husband, Tony
    (Bruce Greenwood), a pharmaceutical kingpin. When Tony takes her
    on a private cruise from Greece to Italy with two other couples,
    Amber is unimpressed by this impromptu no-frills vacation and
    takes out her anger on the ship's first mate, Giuseppe Esposito
    (Adriano Giannini). When a storm leaves the two shipwrecked on a
    deserted island, however, the tables suddenly turn, with
    Giuseppe gaining the upper hand, followed by the two falling in
    love.

  Question: Who does the woman become dependent upon for survival?
\end{lstlisting}
\end{minipage}

\noindent\begin{minipage}{\linewidth}
\textbf{RACE.}
\begin{lstlisting}[style=promptstyle]
  Context: It's eight o'clock and the time for Morning New
    Highlights . One person was killed and two were injured in a car
    accident just thirty minutes ago. Cause of the accident is not
    yet found out. A great fire swept through a storehouse next to
    the Fulton River early this morning, causing about $3,000,000 in
    damage. The fire which broke out at about 2:30 was finally put
    out at 5: 30. That was the Morning News Highlights, Dan Verne
    reporting.

  Question: How many pieces of news are reported in the short
    passage?
\end{lstlisting}
\end{minipage}

\noindent\begin{minipage}{\linewidth}
\textbf{RelationExtraction.}
\begin{lstlisting}[style=promptstyle]
  Context: Q150 was celebrated in 2009.

  Question: When did Q150 occur?
\end{lstlisting}
\end{minipage}

\noindent\begin{minipage}{\linewidth}
\textbf{TextbookQA.}
\begin{lstlisting}[style=promptstyle]
  Context: Above the stratosphere is the mesosphere. Temperatures in
    the mesosphere decrease with altitude. Because there are few gas
    molecules in the mesosphere to absorb the Suns radiation, the
    heat source is the stratosphere below. The mesosphere is
    extremely cold, especially at its top, about -90o C (-130o F).
    The air in the mesosphere has extremely low density: 99.9% of
    the mass of the atmosphere is below the mesosphere. As a result,
    air pressure is very low (Figure 1.1). A person traveling
    through the mesosphere would experience severe burns from
    ultraviolet light since the ozone layer, which provides UV
    protection, is in the stratosphere below. There would be almost
    no oxygen for breathing. And, of course, your blood would boil
    at normal body temperature. Click image to the left or use the
    URL below. URL:

  Question: what is the heat source for the mesosphere?
\end{lstlisting}
\end{minipage}

The self-score prompt is the answering prompt with one field added to the schema and one rule describing it. The question and context are unchanged.

\noindent\begin{minipage}{\linewidth}
\begin{lstlisting}[style=promptstyle]
SYSTEM
  Answer the question using only the given context.
  Reply with a JSON object:
    {"answer": ..., "detailed_answer": ..., "score": ...}
  "answer" must be the shortest exact span from the context that answers the
    question -- no full sentence, no extra words.
  "detailed_answer" is one short explanatory sentence.
  "score" is a number from 0 to 1 giving the probability that your own
    answer is correct. Use a low score when the context or options do not
    clearly support your answer.

USER
  Context: Orange County is a rapidly developing business center that
    includes Downtown Santa Ana, the South Coast Metro and Newport Center
    districts; as well as the Irvine business centers of The Irvine
    Spectrum, West Irvine, and international corporations headquartered at
    the University of California, Irvine. West Irvine includes the Irvine
    Tech Center and Jamboree Business Parks.

  Question: Where are international corporations headquartered?
\end{lstlisting}
\end{minipage}

The LLM-as-Judge prompt is a separate call that contains the same context and question, plus the answer being graded, and tells the judge it does not know the correct answer. The example below grades the answer the model gave to the question above.

\noindent\begin{minipage}{\linewidth}
\begin{lstlisting}[style=promptstyle]
SYSTEM
  You are grading another model's answer to the question below. You do NOT
    know the correct answer. Judge only from what you are given whether the
    answer is correct.
  Reply with JSON: {"reason": "...", "probability_correct": ...}
  "reason" is one short sentence. "probability_correct" is the probability,
    from 0 to 1, that the answer is correct -- 0.0 if it is certainly wrong,
    1.0 if it is certainly right, and a value in between when what you are
    given does not settle it. An empty or evasive answer is not correct.

USER
  Context: Orange County is a rapidly developing business center that
    includes Downtown Santa Ana, the South Coast Metro and Newport Center
    districts; as well as the Irvine business centers of The Irvine
    Spectrum, West Irvine, and international corporations headquartered at
    the University of California, Irvine. West Irvine includes the Irvine
    Tech Center and Jamboree Business Parks.

  Question: Where are international corporations headquartered?

  Answer given: The University of California, Irvine
\end{lstlisting}
\end{minipage}

\clearpage
\end{document}